\documentclass{sig-alternate-2013}
\newcommand{\comment}[1]{}
\usepackage{framed}
\usepackage[show]{chato-notes}
\usepackage{multirow}
\usepackage{graphics}
\usepackage{url}
\usepackage{wrapfig}
\usepackage{algorithmic} 
\usepackage{algorithm} 
\usepackage{float} 
\usepackage{amsmath}
\usepackage{amssymb}
\usepackage{amsfonts}

\numberwithin{equation}{section}

\usepackage{subfig, float, epsfig, epstopdf, balance, algorithm}
\usepackage{helvet}
\usepackage{courier}
\usepackage{multirow}
\usepackage{mathtools}

\newfont{\mycrnotice}{ptmr8t at 7pt}
\newfont{\myconfname}{ptmri8t at 7pt}
\let\crnotice\mycrnotice%
\let\confname\myconfname%

\permission{Permission to make digital or hard copies of all or part of this work for personal or classroom use is granted without fee provided that copies are not made or distributed for profit or commercial advantage and that copies bear this notice and the full citation on the first page. Copyrights for components of this work owned by others than ACM must be honored. Abstracting with credit is permitted. To copy otherwise, or republish, to post on servers or to redistribute to lists, requires prior specific permission and/or a fee. Request permissions from permissions@acm.org.}
\conferenceinfo{CIKM'14,}{November 3--7, 2014, Shanghai, China.}
\copyrightetc{Copyright 2014 ACM \the\acmcopyr}
\crdata{978-1-4503-2598-1/14/11\ ...\$15.00.\\
http://dx.doi.org/10.1145/2661829.2662018}

\begin{document}

\title{How Many Folders Do You Really Need?\\
Classifying Email into a Handful of Categories}

\numberofauthors{1}
\author{
\alignauthor
Mihajlo Grbovic$^1$, Guy Halawi$^2$, Zohar Karnin$^2$, Yoelle Maarek$^2$\\
\affaddr{Yahoo Labs}\\
\affaddr{\{mihajlo, ghalawi, zkarnin\}@yahoo-inc.com}, yoelle@ymail.com\\
\affaddr{$^1$900 First Ave, Sunnyvale CA, USA}\\
\affaddr{$^2$MATAM, Haifa 31905, Israel}\\
}

\date{\nonumber}

\maketitle
\begin{abstract}
Email classification is still a mostly manual task. Consequently, most Web mail users never define a single folder.  Recently however, automatic classification offering the same categories to all users has started to appear in some Web mail clients, such as AOL or Gmail.  We adopt this approach, rather than previous (unsuccessful) personalized approaches because of the change in the nature of consumer email traffic, which is now dominated 
by (non-spam) machine-generated email. We propose here a novel approach for (1) automatically distinguishing between personal and machine-generated email and (2) classifying messages into latent categories, without requiring users to have defined any folder. We report how we have discovered that a set of 6 ``latent'' categories (one for human- and the others for machine-generated messages) can explain a significant portion of email traffic. We describe in details the steps involved in building a Web-scale email categorization system, from the collection of ground-truth labels, the selection of features to the training of models. Experimental evaluation was performed on more than $500$ billion messages received during a period of six months by users of Yahoo mail service, who elected to be part of such research studies. Our system achieved precision and recall rates close to $90\%$ and the latent categories we discovered were shown to cover $70\%$ of both email traffic and email search queries. We believe that these results pave the way for a change of approach in the Web mail industry, and could support the invention of new large-scale email discovery paradigms that had not been possible before.
\end{abstract}
\vspace{-0.1in}
\category{H.4.3}{Information Systems Applications}{Communications Applications}[Electronic Email]
\vspace{-0.1in}
\terms{Information Systems; Algorithms; Experimentation}
\vspace{-0.1in}
\keywords{Email Classification; Machine-generated Email; LDA} 

\section{Introduction}
\label{sec:introduction}
Email classification has been a user's pain point since the early days of email systems. In the last decade, most attempts at automating this task have consisted in mimicking each individual user's personal classification habits~\cite{klimt2004enron}. This highly personalized approach exhibits two main weaknesses: it relies on small datasets as each individual inbox is analyzed independently, and it requires from users to have defined meaningful folders in the first place.  We believe that relying on user-defined folders is the reason for which such methods were never widely adopted. It was shown in~\cite{koren2011automatically} that $70$\% of users do not create even a single folder. Furthermore, we have verified on a large dataset, generated from Yahoo mail,that only $10\%$ of users are ``active classifiers'', the latter being defined as users who moved more than $10$ messages into folders over a period of $6$ months. 

This fact has not remained unnoticed. A few of commercial Web services have changed their approach towards email classification by offering ``classes''  (implemented as labels, or categories for instance) that are common to all users. Thus, AOL mail, with its ``Bulk Senders'' category, or Gmail, with its inbox tabs\footnote{These tabs include  ``Promotions'', ``Social'', ``Updates'' and ``Forums'', see \small \url{https://support.google.com/mail/answer/3055016?hl=en&p=inboxtabs&rd=1}}, offer a few ``common'' classes. On the other end of the range, Koren et al. in~\cite{koren2011automatically}, have analyzed popular folders, in order to identify not just a few but several thousand possible common classes or ``tags''. While the wide range of tags is attractive, one weakness of this approach is the high variance in their level of abstraction. Highly subjective tags such as {\em mom} co-exist with more objective ones like {\em recipes}, or sender-based tags such as  {\em amazon} overlap with related ones like {\em shopping}. In addition, they achieve recall and precision rates of only $80\%$ for a coverage of $72\%$.
This work explores a middle ground direction, where we propose, like in ~\cite{koren2011automatically}, to derive high level-classes from mail data, and especially folders data provided by the minority of users who do define folders. However, similarly to AOL or Gmail, we give preference to fewer, consistent classes, for a more user-friendly experience, and offer these classes to all users, including those who did not define any folder.

First, we propose to exploit the fact that today's commercial Web mail traffic is dominated by machine-generated messages, originating from mass senders, such as social networks, e-commerce sites, travel operators etc.  In their experiments, Ailon et al. stated that machine-generated mail represent more than $60\%$ of Yahoo mail traffic, ~\cite{ailon.2013}.
We argue here that, given the difference in syntax and semantics between human- and machine-generated messages, one should first distinguish between machine and human-generated messages before attempting any finer classification.  Once focusing on machine-generated messages, with huge numbers (millions in some cases) of users receiving the same type of machine-generated messages, data will be much less sparse, and classification techniques should become more precise. In order to validate our intuition, we estimated the actual volume of machine-generated messages on a very large Yahoo mail dataset. This dataset, built for the purpose of this work, covers 6 months of email traffic and more than $500$ billion messages. For privacy reasons, it includes messages belonging exclusively to users who voluntarily opted-in for such studies and message bodies were not inspected by humans. We conducted experiments on this dataset (as detailed later on) and verified that non-spam machine-generated  messages actually represent  $90\%$ of the entire dataset, which encouraged us to proceed.

In the same dataset, we considered the subset of folders data  that include messages classified by users into folders as well as folders labels, generated by more than $40$ million users.  Instead of a priori fixing high level classes, or deriving them from popular folders labels, as done by the previously mentioned efforts, we attempt here another approach. Namely, we propose to discover these categories by applying an LDA (Latent Dirichlet Allocation) approach~\cite{blei2003latent}, on the folder dataset. We explain how ``latent'' categories can be identified in Section~\ref{sec:LDA_cat}. More specifically, we detail how one {\em human} and five machine-generated categories can be inferred directly from the data. We introduce in Section~\ref{sec:methodology} our email classification model and its associated features, such as content, sender, behavioral, and temporal behavioral features. We also justify the different aggregation levels we considered, namely message-, sender- and domain-level aggregation.  An additional challenge is to obtain a sufficiently large training set. Due to privacy issues, the cost of editorial work, and the skewness of the email data, we use various automated methods. For the machine latent categories, our main data source is the folder data. Using our LDA analysis we obtain for a large number of messages a clear linkage to our latent categories. For the human category, folder data is not effective, and we use instead various heuristics leveraging sender information and extrapolated data via \emph{co-training} inspired techniques. We detail how we generated the training data in Section~\ref{sec:training-set}.  Section~\ref{sec:system} describes our classification mechanism and Section~\ref{sec:experiment} presents our results. 
In section~\ref{sec:results} we provide some additional statistics to demonstrate the impact of our classification system.
Note that we exclusively consider consumer Web mail as opposed to corporate email, whose traffic is drastically different in size and nature. Consequently, we did not compare our results to the Enron dataset.  Related work is discussed as relevant throughout the paper.

The key contributions of our work are the following:
(1) we provide new insights on the importance of distinguishing between human- and machine-generated email, (2) we give to the best of our knowledge, the first large scale data-driven validation of the intuition that a few consistent automatically discovered categories cover most of today's Web email traffic, and
(3) we describe an end-to-end Web mail categorization solution, including detailed models, learning mechanisms,  evaluation methods, and results on real traffic.

\section{Discovering Latent Categories}
\label{sec:LDA_cat}
\begin{table*}[t]
  \centering
  {\scriptsize
  \caption{LDA latent topics/categories}
    \begin{tabular}{l|l|l|l|l|l}
     {\bf human} & {\bf career} & {\bf shopping} & {\bf travel} & {\bf finance} & {\bf social}  \\
      \hline
      read 0.06 &  job		.022 & shipping	.016 & travel	 .010 & information	.010 & post	.024  \\
      today 0.04 & jobs		.014 & items	.009 & hotel	.008 & bank	.009  & comment	.021 \\
      great 0.04 & apply	.009 & deals	.008 & deals	.006 & card	.009 & messages	.018  \\
      much 0.04 & search	.009 & sale	.007 & per		.005 & payment	 .008 &photo	.018 \\
      want 0.04 & career	.008 & order	.007 & book	.005 & service	.007 & invited	.017 \\ 
      wish 0.04 & exp		.007 & purchases	.007 & flights	.005 & credit	.007 & attention	.017  \\
     make 0.03 &  national	.006 & shop	.006 & details	.004 & contact	.006 & twitter	.012 \\
     just 0.03 &  capital	.006 & customer	.006 & travelzoo	.004 & customer	.006 & shared	.011 \\
    think 0.03 &   recruitment	.006 & offer	.006 & stay	.004 & security	.005 & replied	.010 \\
    very 0.03 &   manager	.006 & subject	.006 & hotels	.004 & chase	.004 & followers	.010 \\
     \hline
    \end{tabular}
  \label{tab:ldaresults}
}
\end{table*}

Following ~\cite{sebastiani2002}, we define email categorization as the task of {\em ``assigning a Boolean value to each pair  $(d_j,c_i) \in D \times C$, where $D$ is a domain of documents''} (here mail messages) {\em ``and $C$ is a set of predefined categories.''}  Previous research work on personalized email classification used one  $C$ set per user and populated it by reusing the user's existing folders, see \cite{provost1999naive,  brutlag2000challenges, alberts2012email, kiritchenko2011email}. The approach we advocate for here is to adopt one single $C$ for all users. Still $C$ can be defined in many different manners: categories can relate to message importance (see Gmail {\em Priority Inbox}\footnote{\small\url{http://gmailblog.blogspot.co.il/2011/03/new-in-gmail-labs-smart-labels.html}}),  number of recipients (see AOL {\em Bulk Senders}), senders (see popular folders such as {\em Amazon, Facebook}), or types (see Gmail Labs Smart Labels$^2$ {\em Social, Promotion, Updates, Forums, Travel, Finance}) for instance.  In addition, $C$ should be derived from data (rather than predefined) and should fulfill the following requirements.
\begin{enumerate}
\item The number of categories should be small so as not to overwhelm the user. Indeed, it has been shown that, with more than $20$ folders, folder-based discovery becomes less effective than search~\cite{balter2000keystroke}.
\item Categories must be easily interpretable and at the same level of abstraction.
\item Categories should cover a significant amount of email volume.
\end{enumerate}

We first examined all messages that have the potential to be classified by retrieving the most ``popular'' folders in terms of users's ``foldering'' actions. We ignored system folders (e.g.,  ``trash'', ``spam'') and folders created by third-party  services such as OtherInbox\footnote{\small \url{http://www.otherinbox.com}}. We also removed small folders that count less than $1{,}000$ messages, since they would not cover common needs. \comment{, and are  too small for LDA effectiveness~\cite{yan2013biterm}.}  We remained with $100,000$ human-generated folders.
For illustration, we list here the top folders thus identified:  {\em ebay, accounts, personal, school, saved mail, jobs, amazon, taxes, recipes, business, college, bills, house ,facebook, paypal, save, food, linkedin, pictures, surveys, travel, jobs, misc.}  etc.

We then associated with each of these $100,000$ folders an artificial document obtained by concatenating across all relevant inboxes, all messages classified into this folder. We applied LDA~\cite{blei2003latent}, to these  ``document folders''. LDA is commonly used in natural language processing in order to discover a set of latent topics that generated a given set of documents.  In our context, we hoped that latent topics would map into ``latent categories'' , which would then define our set $C$ for further classification. To this effect, we trained an online LDA model~\cite{hoffman2010online}.
\comment{as illustrated in Figure~\ref{fig:lda_alg}}
Recall that LDA receives as input a set of documents (folders), each associated with a bag of words (words of messages put in the folder), and a number of topics $K$. It outputs for each topic the top words associated with it and the topic mixture of each document. We iterated over several values of number of topics, from $K=50$ to $K=5$. To choose the right $K$, we looked at coverage, in terms of portion of traffic covered by folders that were dominantly ($>50\%$) associated with topics. For a value of $K$ and a specific topic, the coverage of the topic is defined as the amount of traffic covered by folders that are associated with said topic by at least $50\%$ by the LDA output. The total coverage associated with a value $K$ is the sum of coverages associated with the topics. Our objective was to find a value of $K$ that would ensure that each individual topic as well as the overall set of topics achieve significant coverage. 
We observed that larger values of $K$ produced topics that were narrow, e.g. {\em Knitting, Cooking, Astrology}  with small coverage. The traffic coverage at $K=50$ was $81\%$, and went down to $74\%$ at $K=25$, to about $70\%$ at $K=6$. At $K=6$ the coverages of the topics were sufficiently large (see Table~\ref{tab:traffic_results}), while at larger values of $K$ narrow topics emerged.

We further examined the topics obtained for $K=6$, as this value exposed a good balance between total and individual coverage, 
by inspecting their associated vector of words and weights. As evidenced in  the five rightmost columns of Table~\ref{tab:ldaresults}, the words associated with each of these fives topics clearly shared a common underlying concept, respectively {\em career, shopping, travel, financial} and {\em social}. 
One topic remained that was a bit puzzling at first, as it contained many highly frequent (stop) words and no key concept seemed to emerge from them. After some quick  examination of the messages (simply looking at senders) associated with this topic, we discovered that most were human-generated, as opposed to the  five other topics that are mostly machine generated. The word associated with this first topic are quite frequent in personal communications, consequently we  named this last topic {\em human}. \comment{which also reminds of Gmail Inbox  {\em Primary}  tab that includes messages from friends and family in addition to the messages not covered by other tabs. }

Given these 6 labeled topics, we further verified their distribution over a sample of popular folders, as shown in Table~\ref{tab:foldertopics}. Some mappings between folders and latent topics were almost perfect, like the folder ``financials'' being associated with the latent topic {\em financial} with a $96\%$ weight, or ``hotels'' with the topics {\em travel} at $74\%$, with still a flavor of {\em financial} at $15\%$. More interestingly, a highly subjective folder like ``mom and dad'' had no clear winner. While it did reflect the {\em human} topic at $31\%$, it also (and ironically so) had {\em financial} at $55\%$!  Based on this analysis, we finalized our decision of choosing as $C$ $\{${\em human, career, shopping, travel, financial}$\}$, as it meets our 3 requirements of small size, same abstraction level and coverage.

\comment{
\begin{figure}[t]
\centering
{\includegraphics[width=0.45\textwidth]{folder_fig.eps}} 
\caption{Distribution of messages and senders over folders ($6$ months)}
\label{fig:folder_hist}
\end{figure}}

\comment{
\begin{table}[t]
\centering
{\scriptsize
  \caption{Top folders ordered by volumes (top to bottom, then left to right)}
    \begin{tabular}{l | l  | l}
      {\bf folder name} & {\bf folder name} & {\bf folder name}\\
      \hline
unsubscribe & archive & accounts \\
ebay & school & saved mail  \\
personal & amazon	& taxes   \\
jobs	& business & college    \\
recipes & house & facebook	 \\
bills & save & food	 	  \\
paypal & pictures	& surveys 	 \\
linkedin & job search &miscellaneous	  \\
travel	& notes 	& jokes \\
family & stuff & pch	 \\
orders& groupon& chase \\
shopping & purchases	& pinterest	 \\
friends & health & online orders	 \\
misc & other & craigslist		 \\
saved & job & real estate \\
receipts	& bank& banking	 \\
coupons & house  & mom \\
work & church & insurance	 \\
untitled & keep & confirmations \\
important & wedding	& payments	  \\
2013	& finance 	 & compras\\
    \end{tabular}
  \label{tab:resultstable}
\end{table}}
}

\begin{table}[t]
\centering
{\scriptsize
  \caption{Folder labels (LDA assigned)}
    \begin{tabular}{l | l}
      {\bf folder name} & {\bf topic distribution} \\
      \hline
financials &  $96\%$ financial \\
mom and dad & $55\%$ financial, $31\%$ human, $18\%$ shopping \\
linkedin & $56\%$ social, $34\%$ career \\
jobs & $89\%$ career, $19\%$ financial \\
hotels & $74\%$ travel, $15\%$ financial \\
bestbuy & $76\%$ shopping, $21\%$ financial\\
church & $38\%$ financial, $23\%$ human,  $8\%$ shopping \\
aaa.com & $59\%$ travel, $35\%$ financial \\
ebay & $96\%$ shopping 
    \end{tabular}
  \label{tab:foldertopics}
}
\end{table}

\section{Modeling Email}\label{sec:methodology}
One key challenge in any classification task is data modeling and fixing the level of granularity of data points i.e., the objects to classify.  One option is to consider each individual message as a single data point, associated with various features extracted from the message header and body. A second approach consists in aggregating messages at higher levels, for instance at the SMTP address level~\cite{SMTP}, (referred to as ``sender''), or at the mail domain level (easily extracted from the SMTP address by truncating its prefix up to the character ``@''.)  This latter approach not only helps with sparseness issues but more importantly allows for much more efficient processing, without requiring expensive body analysis and feature extraction. This is critical at the scale of  Web mail, when huge numbers of messages need to be classified at delivery time. We explain how we use a combination of both approaches in Section~\ref {sec:system}. Feature extraction and different aggregation levels are described in the rest of this Section.

\comment{For the purposes of this paper we looked at the portion of Yahoo Email users that opted-in using their data for purposes of email categorization. We extracted incoming and ongoing messages from these email addresses during a $6$ month period. All the statistics mentioned throughout the paper are based on this snapshot of data.}

\subsection{Extracting Features}
\label{sec:features}
We follow the general categorization framework where numerical features are extracted from each object (in our case, message, sender or domain) later to be fed to a learning mechanism that given labelled examples, e.g.\ messages with manually assigned categories, will be able to categorize unlabeled objects. In this section we describe the types of features and how they are obtained.

A data point 
consists of content features, address features, and behavioral features which include a special subcategory of temporal behavior features. 

{\bf Content features} include features derived from the message subject and body. We extract words from the subject line and message body, as well as the subject character length, body character length and the number of urls occurring in the body. To form sender features, we use the total counts of observed words across all messages from that sender, as well as the average, the minimum and the maximum subject and body lengths and url counts. In large databases, such as the one we are dealing with, the most frequent words can easily occur millions of times (e.g. ``the'', ``a'', ``in'').  We eliminate the top $400$ words found across most senders, as well as the usual stop words\footnote{\small \url{http://anoncvs.postgresql.org/cvsweb.cgi/pgsql/src/backend/snowball/stopwords}} for several common languages. Finally, we filtered out the low frequency words that occur in less than $100$ different senders.

{\bf Address features} include features extracted from the sender email address, including the subdomain, e.g. (.edu, .gov, etc.), subname (e.g. billing, noreply). To extract the subdomains we split the domain part of the address (after the ``@'' character) at the delimiter ``.''  location, and use the resulting words as features. We do the same for the subnames, but on the name portion of the email (before the ``@'' sign). We also extract ``commercial'' keyword from senders. To this effect, we use a list of more than $500$ keywords, typically used as bid words in ad targeting, such as {\em itinerary, flight, ticket, order, confirmation, billing, payslip, payment, transaction, stocktrade, career, shopping, travel,} etc., and look for a match in the email address. To counter the imbalance between rare and frequent words, we remove the $800$ most frequent sender substrings. This avoids poor generalization when using biased training data. For example, very common domain names, such as ``yahoo.com'', or ``gmail.com'', could get picked as the most informative features in a model trained to detects human senders. This rule clearly does not generalize well outside the training dataset.
In addition to removing the most frequent substrings, we remove the substrings that appear in less than $100$ senders. We also map very rare or random strings into a canonical form, for instance \texttt{5tsdfocfyf66c@bookstore.com}, is mapped into a single feature ``random string''.

{\bf Behavioral features} include features extracted from the sender's and recipient's actions over a given message. They can be split into {\it outbound}, {\it inbound} and {\it action} features. The {\it outbound} features for the $n$-th sender cover the sender's outgoing activity, such as weekly and monthly volumes of sent messages, histogram of the number of recipients in their messages, volumes of messages sent as a reply (indicated by ``re:'' in the subject line), volumes of messages sent as forward (with FW: in the subject line). The {\it inbound} features for the $n$-th sender cover the sender's incoming activity, such as volume of the messages received by the sender, as regular, reply or forward messages. 
The {\it action} features for the $n$-th sender cover the activity of the sender's email recipients, such as how many times the messages from the $n$-th sender were read, deleted, replied to, forwarded, ended up in spam or any other folder. To create these features, raw counts are converted to percentages of total inbound volume (e.g., percentage of $n$-th sender messages moved to trash). The names of the  folders to which a given sender messages have been moved, are especially interesting: they can be seen as a human label of a specific sender. Therefore, for the $n$-th sender, we also use the names of folders (with the counts of moves to that folder ) as folder features. We remove the folders that contain less than $100$ different senders, as well as the $50$ folders with the highest number of senders, including system folders (``trash'') and third-party services folders.

{\bf Temporal behavior features} form a subcategory of the behavioral features. They reflect the frequency of specific actions over a given period of time. For instance, we record whether a sender sends more than $X$ messages in an hour. These features are represented as a histogram, where $X$ takes as values: $10$, $60$, $80$, $100$, $120$. We refer to them as the {\it burst} features. 

Many of the features described above are counts of actions or words. To handle large counts, we normalize the data at an instance level by using $\log(1+x(k))$ instead of the original $k$-th feature value $x(k)$. Features that represent percentages of total traffic (e.g. moves to trash, spam, etc.) are left in the $0$ to $1$ range.

\subsection{Email Aggregation Levels}
\label{sec:aggregation}

\begin{table}[t]
\centering
  {\scriptsize
  \caption{Generalization loss in domain-level categorization}
    \begin{tabular}{l|ll}
      {\bf domain} & {\bf senders} & {\bf category} \\
      \hline
      \multirow{4}{*}{\scriptsize{vailresorts.com}} &store@vailresorts.com & shopping \\
      & careers@vailresorts.com & career  \\
      & booking@vailresorts.com & travel  \\
      & payroll@vailresorts.com & financial \\
      \hline
      \multirow{4}{*}{\scriptsize{asmnet.com}} & payroll.services@asmnet.com & financial  \\
      & payrollquestions@asmnet.com & financial  \\
      & .*.services@asmnet.com & shopping \\
      & careers@asmnet.com & career \\
      \hline
      \multirow{5}{*}{\scriptsize{target.com}} & orders@target.com & shopping \\
      & target.jobs@target.com & career  \\
      & target.payroll@target.com & financial  \\
      & name.lastname@target.com & human \\
      & .*.service@target.com & shopping \\
      \hline
      \multirow{4}{*}{\scriptsize{idfc.com}} & investormf@idfc.com & financial \\
      & customerservice@idfc.com & financial  \\
      & name.lastname@idfc.com & human \\
      & traveldesk@idfc.com & travel \\
    \hline
     \multirow{4}{*}{\scriptsize{meilgroup.com}} & payroll@meilgroup.com & financial \\
      & traveldesk@meilgroup.com & travel  \\
      & lastname@meilgroup.com & human \\
      & careers@meilgroup.com & career \\
    \end{tabular}
  \label{tab:examplestable}
}
\end{table}

We consider three levels of aggregation, by message, sender and domain. To choose the right level of aggregation for our system, we consider what we lose and what we gain by aggregating to the next level, starting from the level of the email message. We consider several dimensions: generalization loss,  data size and  classification scalability.

{\bf Generalization loss:}  Aggregating messages offers some risks, if by aggregating we put in the same class messages that should have been mapped to different categories. One typical mistake that occurs when aggregating by sender occurs when a given sender changes its behavior. It happens for instance when a sender (such as a recruiter, finance broker or a travel agent) sends machine generated messages to clients in $90\%$ of the cases,  and personal messages otherwise. We refer to this issue as the {\it machine-human mix}. In some rare cases, we observed another type of loss we call the {\it multiple businesses loss}, where a sender is a company that has more than one type of business and sends all messages from the same address. Although many companies conduct multiple types of business, they typically use different addresses for each type, hence the {\it multiple businesses loss} is quite low. Domain-level aggregation, as the most aggressive type, suffers the most loss; in particular the multiple businesses loss is quite common there. Table~\ref{tab:examplestable} shows some examples of generalization loss due to aggregating at a domain level. 

\begin{table}[t]
\centering
{\scriptsize
\caption{Categorization level data comparison}
\begin{tabular}{l |ll}
 {\bf type} & {\bf \#examples} & {\bf avg \#features}  \\
\hline
message-level & $1.4$T & $6.31$  \\
sender-level (raw) & $2.6$B & $26.86$  \\
sender-level (filtered) & $339$M &  $29.22$  \\
domain-level & $75.8$M & $34.61$ \\
\end{tabular}
\label{tab:catlevels}}
\end{table}

{\bf Data size:} Data size is a key factor in deciding the right aggregation level as learning and categorizing at this scale is most challenging. Table~\ref{tab:catlevels} shows the number of data points in our  dataset for the various aggregation levels.  As expected sender-level aggregation drastically reduces the data size, going from more than one trillion messages to about $330$ million senders after aggressive filtering. Domain-level aggregation brings an additional decrease, bringing us to about $75$ million domains, a significant drop but not as drastic as the one brought by sender-level aggregation. 

{\bf Classification scalability:}  The highest gain in scalability is between sender and message aggregation. Indeed, extracting features from each message and running a classifier upon delivery might be extremely costly. On the other hand, mapping each message into its sender presents huge performance advantages. The classification process can be conducted offline on all the previously seen senders, and an incoming message can be assigned the category associated with its sender via a simple (yet very large) table lookup.

Consequently, we propose here a two-stage approach. The majority of the email traffic should be classified by sender for performance optimization. However, whenever the classifier returns a low confidence score for a given message sender, the more costly message-based classification should be applied. This approach is detailed in Section~\ref{sec:system}.

\section{Training Data}
\label{sec:training-set}
One key challenge in Web mail classification is to generate a labeled training dataset given the inherent size of the data as well as  privacy issues. We consider here 3 types of labeling techniques: manual, heuristic-based and automatic, which  we discuss below. We used as labels our 6 latent categories, as well as the {\em human} and {\em machine} labels to differentiate between human- and machine-generated messages.

\subsection{Manual labeling}
Manual labeling consists of having human editors assign labels to specific examples. This method has some clear weaknesses. Given the data sensitivity, even with users that agreed having their mail examined, only a limited number of responsible editors can be involved. Then it becomes more difficult to have multiple editors looking at the same data, which is a must given the nature of the task. Indeed, looking at a confusion matrix, we observed many inconsistencies among editors, for instance between {\em Shopping} and {\em Finance} or {\em Travel} and {\em Finance}. We nevertheless generated such a dataset, denoted as  ${D}_{\mathrm{man}}$,  by giving to $5$ paid editors a pool of about  $18,000$ messages originating from about $14,000$ senders.  This pool contained multiple messages per sender, with some overlap between editors' assignments. Close to $400$ inconsistent labels were generated, which were resolved via additional editor intervention.

\begin{table*}[t!]
\centering
  {\tiny
  \caption{Folder labels (manually assigned)}
    \begin{tabular}{l|l|l|l|l|l}
      {\bf shopping} & {\bf finance} & {\bf travel} & {\bf social} & {\bf career} & {\bf other}  \\
      \hline
      orders	&bank statements&	hotel reservations&	groups&	jobs&	church \\
purchases	& trading	&hotels	&social	&job applications	&stuff \\
online orders	&credit cards 	&train tickets&	facebook	&recruiters	&soccer \\
receipts	&credit 	&flights	&facebook friends	&job search	&dating \\
shopping	&financial &	hotels - airlines&	twitter	&trabajo	&match.com \\
compras &	bank	 &travel reservations	&myspace	&career	&school\\
internet recipes	&banking	&travel info&	social networking&	resumes	&education\\
ebay	&bank stuff	&hotel	&social network	&job apps	&jokes\\
shoes &	bill confirmations	&travel confirmations&	social networks&	job hunt	&college\\
amazon	&bills to be paid&	air france &	social sites	&job hunting	&surveys\\
\comment{
online shopping	&auto insurance	&voyages	&social groups	&jobs applied for	&astrology\\
online purchases	&health insurance	&voyage	&online communities	&jobs 2013	&health\\
internet shopping	&finance trade	&airlines	&online groups	&job search 2013	&cooking\\
shopping receipts	&billpay	&bus tickets 	&yahoo groups	&jobs applied	&recipes cooking\\
order confirmations	&taxes	&travel 	&book groups	&job application	&dating mail\\
coupons	&2012 tax deductions	&air tickets	& flickrmail &	jobsearch	&hobbies, crafts\\
shopping offers&	real estate	&airline tickets&	tumblr&	recruiters &	astrology,love\\
shopping and rebates&	accounts, bills	&business travel	&instagram emails	&recruitment	&television\\
shopping-misc	 & bills and statements	&vacation reservations	&pinterest	&cv	&titans\\
gifts - bags - etc	&pay slip	&trip confirmations	&netlogmail	&employment	&entertainment \\}
     \hline
    \end{tabular}
  \label{tab:labeledfolders}
}
\end{table*}

\subsection{Heuristic labeling} 
Heuristic labeling consists of applying heuristic rules derived from world knowledge. Such labeling achieves high precision but is limited in coverage. We used this type of labeling mostly for differentiating between human and machine senders.  One key challenge here lies in the fact that SMTP domain information is not sufficient. Indeed, large Web mail services are not only used by humans. Non-spam machine-generated messages can originate from domains such as {\tt gmail.com}, {\tt hotmail.com}, or {\tt yahoo.com}, even for small distribution volumes.  Conversely, domains such as {\tt amazon.com} or {\tt ibm.com, intel.com} or {\tt hp.com} can be associated with both machine and personal communications of corporate employees. We focused on corporate domains as they represent the main source of errors between human and machine labels, with the following simple heuristics.
In order to identify corporate machine senders, we spotted reserved words such as {\tt mailer-daemon} or {\tt no-reply}, or  repeating occurrences  of words such as {\tt unsubscribe} in message headers. We also used signals such as spam votes.  In order to identify human senders, we looked for patterns such as {\tt <first name>.<lastname>} and other various regular expressions. 
We also applied a semi-supervised approach similar to co-training~\cite{blum1998combining},  in order to further increase our heuristic dataset, remove false positives, and most importantly obtain a less skewed sample. In a nutshell, the idea is to split the features into two independent sets, train a classifier on one set, then use the labels with high confidence to train a classifier on the second set, then possibly iterate on all features with all of the labels.  
\comment {Yoelle: did not understand not enough details so i removed - zk please review 
In our case, the initial training set consisted of initial human and machine senders with features based almost solely on the sender address and had a very weak correlation with other features related to the sender behavior. Hence, we trained a model for identifying human vs.\ machine sent email using only the behavioral features and the initial dataset. }
Eventually, a  heuristic dataset denoted ${D}_{\mathrm{h-m}}$ was generated, which consists of $60,000$ human senders and $80,000$ machine senders. Most of the machine senders were further assigned machine latent categories as labels,  using the techniques outlined below. The remaining ones were labeled as {\em Other} in the final dataset used for training of the production model.

\subsection{Automatic labeling} 
For automatic labeling we apply two types of techniques, folder-based majority and LDA voting.

\begin{algorithm}[t!]
{\scriptsize
\caption{\small {Folder-based voting for assigning labels to senders}}
\label{lst:alg1}
{\bf Inputs:} Set of labeled folders $\mathcal{F}$, Unlabeled sender dataset ${D}_u$, vote threshold $\tau_v=50$, folder threshold $\tau_f=2$ \\
{\bf Output:} Labeled sender dataset ${D}_l$ (subset of entire dataset ${D}_u$)
\begin{tabbing}
1. {\bf Initialize} ${\cal D}_l$ to empty set \\
2. {\bf For} \= each unlabeled sender ${\bf x}_n=1,...,N_u$ \\
3. \ \ \  {\bf Find} labeled subset of folders $f_i \in \mathcal{F}$ that sender $n$ \\
 \ \ \ \ \ \ \ was moved to by the user, with counts of moves $c_{f_i}$ \\
4. \ \ \  {\bf Sum} the move counts by labels $y \in \mathcal{Y}$, $c_y=\sum_{f_i \in y} c_{f_i}$ \\ 
5. \ \ \  {\bf Identify} the class with most votes $y_n$ and number of \\
 \ \ \ \ \ \ \ folders that voted for that class $num_{f_i \in y} $ \\
6. \ \ \  {\bf If} $c_y>\tau_v$ and $num_{f_i \in y} > \tau_f$ \\
7. \ \ \ \ \ \ \ \ \ {\bf Label} sender $n$ with class $y_n$, add it to ${D}_l$ \\
8. \ \ \  {\bf End} \\
9. {\bf End} \end{tabbing}
}
\end{algorithm}

{\bf Folder-based majority voting} (see Algorithm~\ref{lst:alg1}) leverages user-generated folders to obtain ground truth labels for senders. There are hundreds of folder names that relate to ``shopping", ``finance", or ``travel", etc. Since there are much fewer folders than there are senders, we manually labeled close to $600$ folders that carry the meaning of our latent categories, and let the folders vote for senders label.  A subset of the labeled folders is shown in Table~\ref{tab:labeledfolders}.  Note that we did not include the {\em Human} category since we hardy observed any folder that would be perfectly aligned with it. Even our previous example of {\em mom and dad} folder in Table~\ref{tab:foldertopics} includes machine-generated messages as evidenced by its strong {\em Finance} topic. We also saw multiple examples of human folders, related to family and friends, that contained messages from known vendors or travel companies. The labeling procedure is detailed in Algorithm $1$, with threshold parameters chosen via grid search.

For each sender, we identify the folders with labels that contain messages originating from this sender. Then, the folders vote for the labels. If there is enough confidence for the winning label, as per threshold, the sender is labeled accordingly. Several examples of the labeled senders with individual folder votes are given in Table~\ref{tab:foldertopicsLDA}. After applying this procedure to the entire dataset of senders, we end up with more than $81,000$ labeled senders that form ${D}_f$. In the process of merging ${D}_f$ with ${D}_{\mathrm{man}}$ and ${D}_{\mathrm{h-m}}$, there were approximately $5,000$ matches (same label assigned) and $700$ conflicts (different label assigned). Senders that were labeled as machine in ${D}_{\mathrm{h-m}}$ and were assigned a machine latent category in ${D}_f$ or ${D}_{\mathrm{man}}$, were not counted as conflicts. The conflicts were resolved by human intervention. We denote the resulting merged dataset by $D_{v1}$.
\begin{table}[h]
\centering
{\footnotesize
  \caption{Labeled dataset ($D_{v2}$)}
    \begin{tabular}{l l | l l}
      {\bf class} & {\bf \# labels} &  {\bf class} & {\bf \# labels} \\
      \hline
      {\bf human} & $68,322$ & {\bf career} & $18,075$ \\
      {\bf shopping} & $42,549$ & {\bf other} & $15,796$ \\
      {\bf travel} & $33,888$ &  {\bf social} & $5,480$ \\
      {\bf financial} & $26,459$ &
    \end{tabular}
  \label{tab:categorysize}
}
\end{table}

{\bf Folder-based LDA voting} is the second technique we used in order to scale-up the majority vote technique. It is needed mostly because the majority of folders cannot be labeled, either because the folder name is not descriptive enough or it contains a mix of messages. As one of the results of LDA training, we know the topic distributions of $100,000$ folders counting more than $1,000$ messages, as illustrated in Table~\ref{tab:foldertopics}. We used these topics for ``soft voting'': instead of assigning $1$ vote, each folder assigns partial votes to the senders of the messages it contains, using topic weights. \comment{see Figure~\ref{fig:folder_hist}. } We then re-normalized the votes to sum up to 1 in order to create a labeled set from the senders with a topic score above $80\%$. This process produced  $42,000$ labeled senders that already existed in $D_{v1}$ as well as $56,000$ additional new senders. The labels of senders that were already in $D_{v1}$ mostly agreed with the existing ones, with only $121$ conflicts, not counting the machine labels that received a subcategory label. The final, merged dataset $D_{v2}$ contains $210,000$ senders. Table~\ref{tab:categorysize} shows the counts of human and machine labels in $D_{v2}$.

\begin{table}[t!]
\centering
{\tiny
  \caption{Examples of folder voting (majority and LDA)}
    \begin{tabular}{l | l}
     {\bf sender} & boots@email.boots.com \\
{\bf folders} & shopping:55 online shopping:6 ebay:4 on-line shopping:4 \\
&amazon:2 finance:1 bills:1 cv:1 orders:1 paypal:1 \\
&internet shopping:1 car insurance:1 receipts:1 credit cards:1 \\
{\bf majority} & shopping \\
{\bf LDA} & shopping \\
\hline
{\bf sender} & 1800usbanks@online.usbank.com \\
{\bf folders} & bank:69 banking:29 bills:13 banks:8 bank statements:5 \\
&finance:5 bank stuff:4 mom:3 purchases:2 receipts:2 \\ 
& girl scouts:2 financial:2 ebay:1 car insurance:1  \\
&online bills:1 online orders:1 credit:1 bill payments:1 \\
{\bf majority} & financial \\
{\bf LDA} & financial \\
\hline
{\bf sender} & accorhotels.reservation@accor.com \\
{\bf folders} & travel:336 voyages:120 hotel:107 hotels:97 \\
&compras:23 receipts:21 hotel reservations:9 flights:9 orders:8\\
&cv:5 shopping:5 travel confirmations:5 air tickets:4 travel\\
&reservations:3 jobs:2 bills:2 flight:2 airlines:2 travail:1 dad:1 \\
{\bf majority} & travel \\
{\bf LDA} & travel 
    \end{tabular}
  \label{tab:foldertopicsLDA}
}
\end{table}

\section{Classification Mechanism}
\label{sec:system}
\begin{figure}[h]
\centering
{\includegraphics[width=0.39\textwidth]{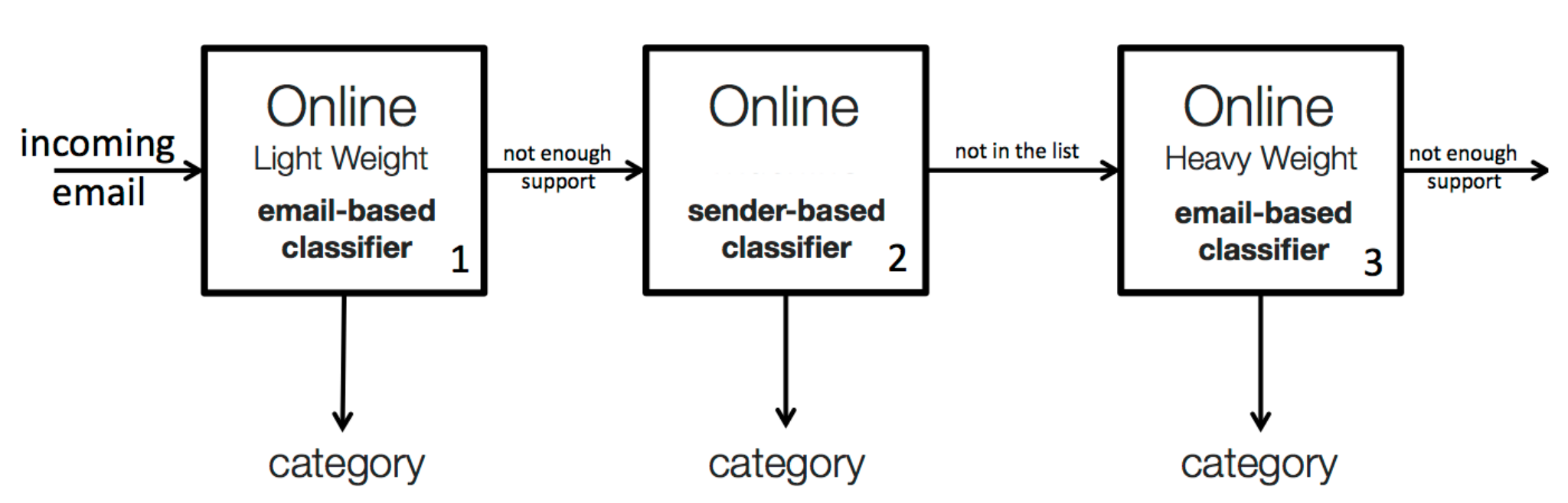}} 
\caption{Overview of the end-to-end system}
\label{fig:overview}
\end{figure}

The classification mechanism put in place has two major elements.  The first is the online mechanism, designed in a cascading manner mainly to account for scalability. The online system leverages both message-level and sender-level classifiers in order to categorize incoming messages, while taking into consideration the strict requirements of a real time Web mail system. The system has three stages: (1) lightweight classification, (2) sender-based classification, and (3) heavy-weight message-based classification.
The second component is the offline component where (1) we periodically classify the known email senders into the 6 categories and (2) train a multi-label message classifier. The online system diagram is given in Figure~\ref{fig:overview}. We describe its components in detail below. 

{\bf Online lightweight classification:} The initial classification is at a message-level, consisting of hard-coded rules designed to quickly classify a significant portion of email traffic. Easy cases include messages from the top 100 senders that cover a significant percentage of the total traffic and are category consistent. Also, as a heuristic decision, we categorized all reply/forward messages as human. By taking precedence over sender-level, message-level classification helps avoid generalization errors due to mixed human-machine senders. As for performance, the process described requires very few resources and covers 32\% of the email traffic.

{\bf Online sender-based classification:} The second phase in our cascade classification process involves looking for the sender in a lookup table containing senders with known categories. This table is created in our offline component, described below. A sender that does not appear in said table is absent due to one of the two reasons; (1) it was either never, or hardly seen in the past meaning that the sender features are too sparse and noisy to be useful, or (2) the offline classification process did not have an answer with a sufficient confidence. The second type of missing senders are those with the most chance of being mixed, meaning that the messages originating from them should not all be assigned the same category. The amount of traffic that is not covered by this phase is roughly 8\%.   

{\bf Online Heavy-weight classification:} Email messages whose sender did not appear in the classified sender table are sent to a heavy-weight message based classifier. As only 8\% of the traffic end up in this last phase we can afford slightly heavier computations than we would have afforded had we employed the classifier on all incoming email. Here, we use all relevant feature, pertaining to the message body, subject line and sender name.

{\bf Offline creation of classified senders table:} We use the training set $D_{v2}$ described in Section~\ref{sec:training-set} in order to train a logistic regression model. The details of the implementation for both training and classification are detailed in Section~\ref{sec:experiment}. For each category we train a separate model in a one-vs-all manner, meaning that the senders assigned any different label are considered negative examples of equal weights. 
The classification process is run performed periodically to account for new senders, and modified behavior patterns (the features of the senders are refreshed at each run) of existing senders. We output for each sender a category and confidence score; if the confidence is sufficiently high then the sender enters the table. 
The process consists of the following: All 6 classifiers (one for each category) are run on all of the senders in our data base. To determine the category of a sender we choose, from the classifiers returning a positive answer, the one with the highest confidence, and set this score as the final confidence. If all of the classifiers returned a negative answer, the chosen category is `other', meaning machine generated that does not fit our predefined categories; the category confidence is set as the lowest confidence score of the individual classifiers. 

{\bf Offline training the message-level classifier:} This task is not done periodically but is done once based on the labeled data described in Section~\ref{sec:training-set}. The training process is quite similar to the sender classification in the sense that a logistic regression model is trained for each category in a one-vs-all model. The major difference is the training set, which is of course different as it contains messages rather than senders. To obtain it, from each sender in $D_{v2}$ we choose $5$ random email messages, and assign them the label of their sender. The features associated with the messages do not include any data on the sender. The other difference when compared to the sender-based classifier is that here we do not allow a non-decision, as this component is used in the final level of the online cascade classifier. Hence, there is no issue of having sufficiently large confidence. Due to space constraints, we do not elaborate on this process further but note that it is quite similar to the sender classification process and has relatively low impact (only 8\% of the traffic is run through this process).

\vspace{-0.1in}

\begin{figure*}[th]
\centering
\subfloat[ROC curve]{\label{fig:roc}\includegraphics[width=0.395\textwidth]{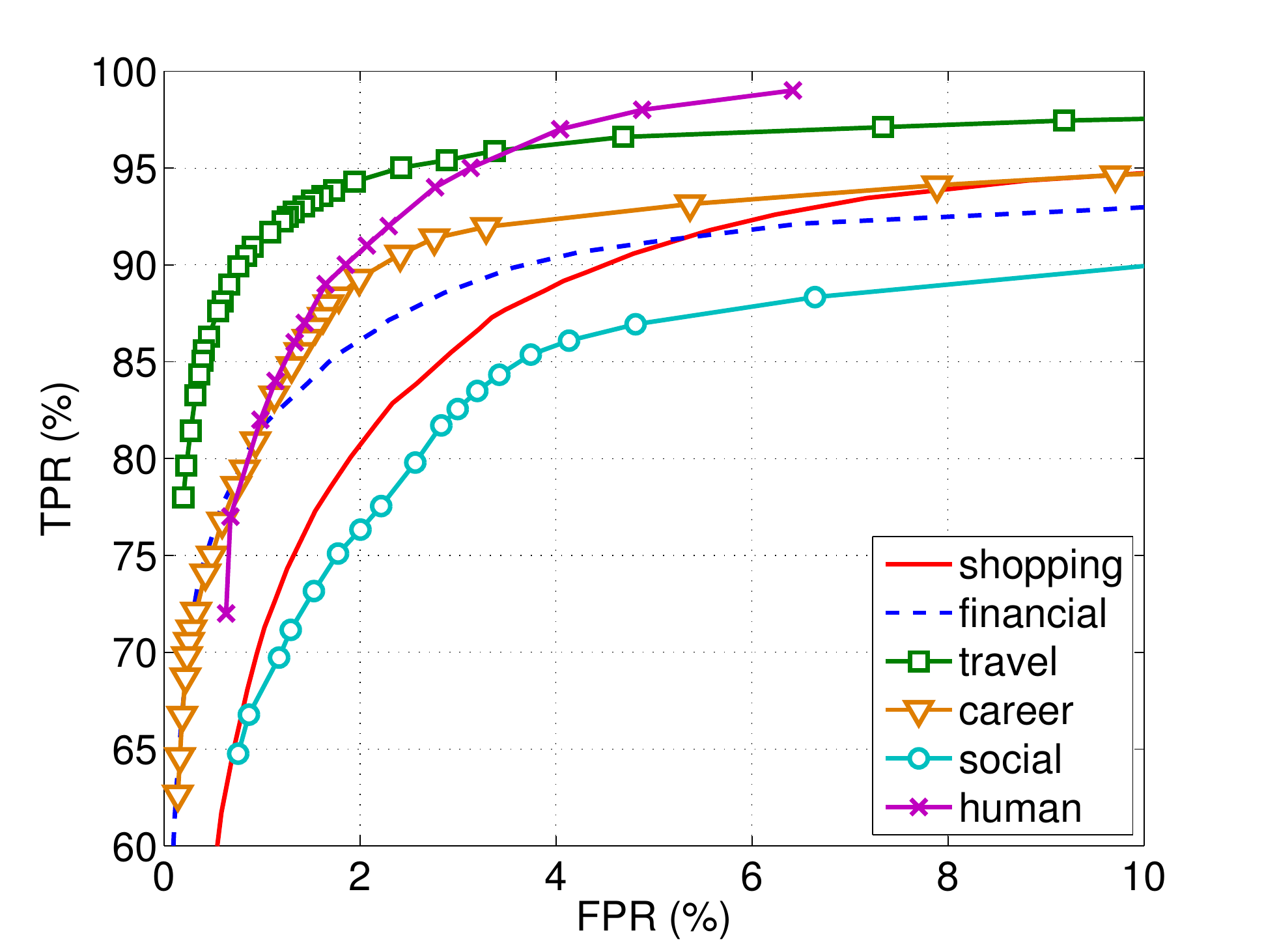}} 
\subfloat[PR curve]{\label{fig:prcurve}\includegraphics[width=0.39\textwidth]{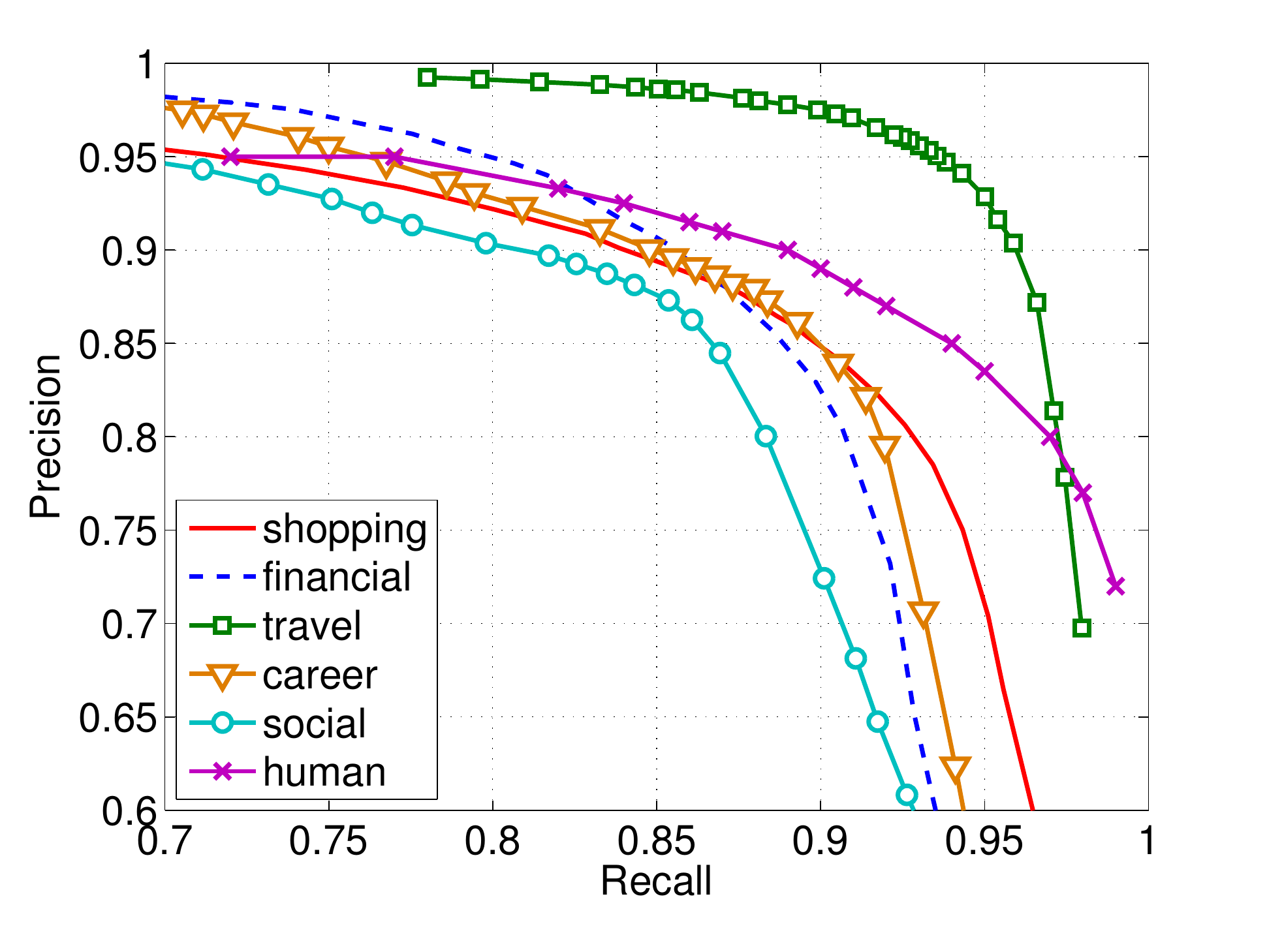}}
\caption{Performance Results}
\label{fig:resultsROC}
\end{figure*}

\begin{table*}[t]
\center
{\scriptsize
  \caption{AUC (one vs rest classification) Performance on different feature subsets}
    \begin{tabular}{l | l | l l l | l l }
      {\bf category} & {\bf all features} & {\bf content only} & {\bf address only} & {\bf behavioral only } & {\bf no burst} & {\bf no body} \\
      \hline
shopping &  $0.971$ & $0.966$ & $0.912$ & $0.645$ & $0.973$ & $0.967$  \\
financial & $0.970$ & $0.967$ & $0.899$ & $0.686$ & $0.968$ & $0.964$ \\
travel & $0.989$ & $0.987$ & $0.932$ & $0.722$ & $0.989$ & $0.986$  \\
career & $0.985$ & $0.976$ & $0.872$ & $0.647$ & $0.977$ & $0.969$ \\
social & $0.936$ & $0.918$ & $0.821$ & $0.614$ & $0.922$ & $0.910$  \\
human & $0.987$ & $0.971$ & $0.948$ & $0.982$ & $0.983$ & $0.987$  \\
\hline
avg & $0.973$ & $0.964$ & $0.897$ & $0.716$ & $0.973$ & $0.971$ \\
    \end{tabular}
  \label{tab:resultsSelective}
}
\end{table*}

\section{Experiments and Evaluation}
\label{sec:experiment}
\begin{figure*}[th]
\centering
\subfloat[{\small Read ratio}]{\label{fig:humanRead}\includegraphics[width=0.25\textwidth]{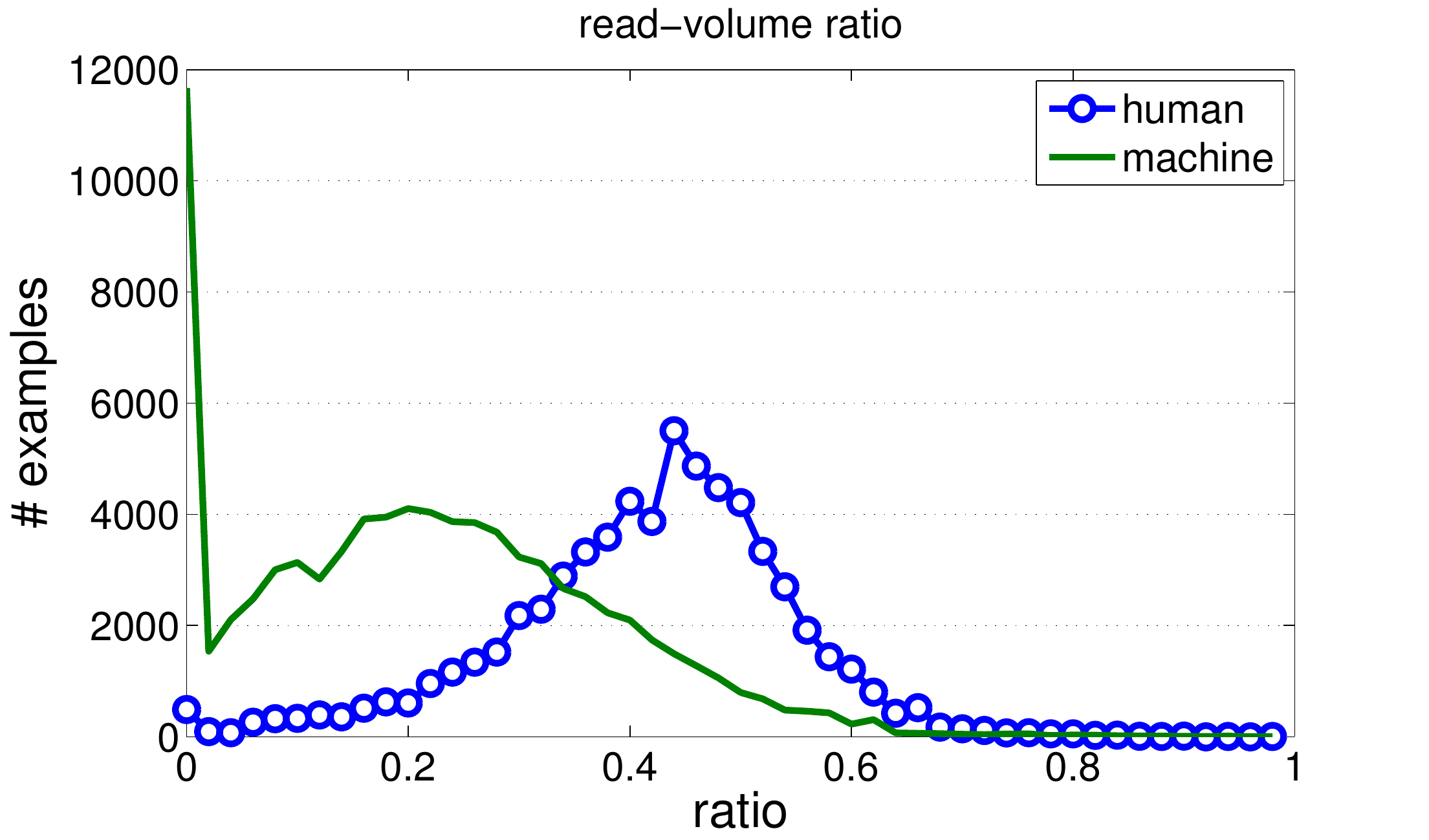}} 
\subfloat[{\small {\tt unsubscribe} occurrence}]{\label{fig:humanUn}\includegraphics[width=0.25\textwidth]{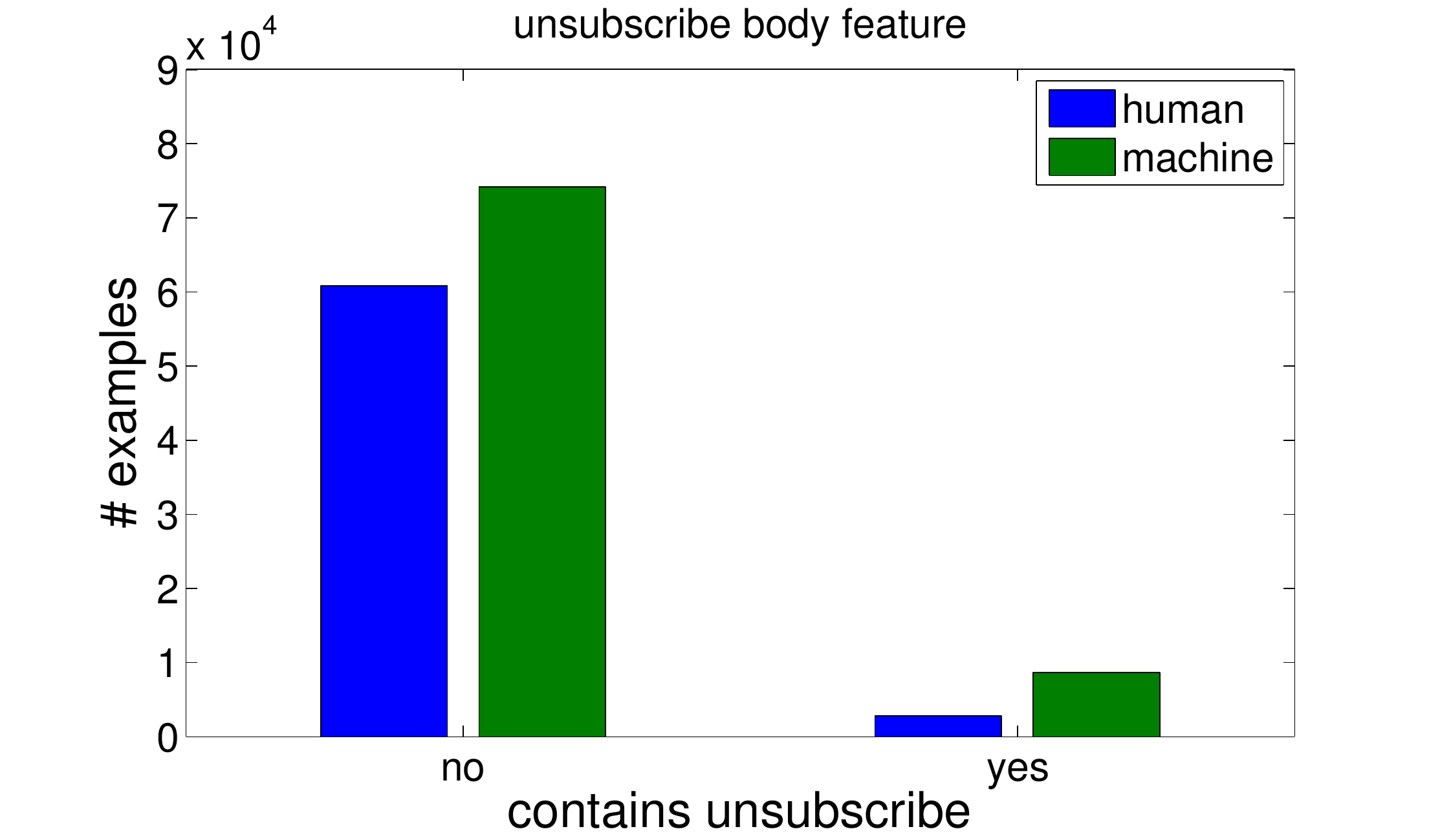}}
\subfloat[{\small Common name occurrence}]{\label{fig:humanCommon}\includegraphics[width=0.25\textwidth]{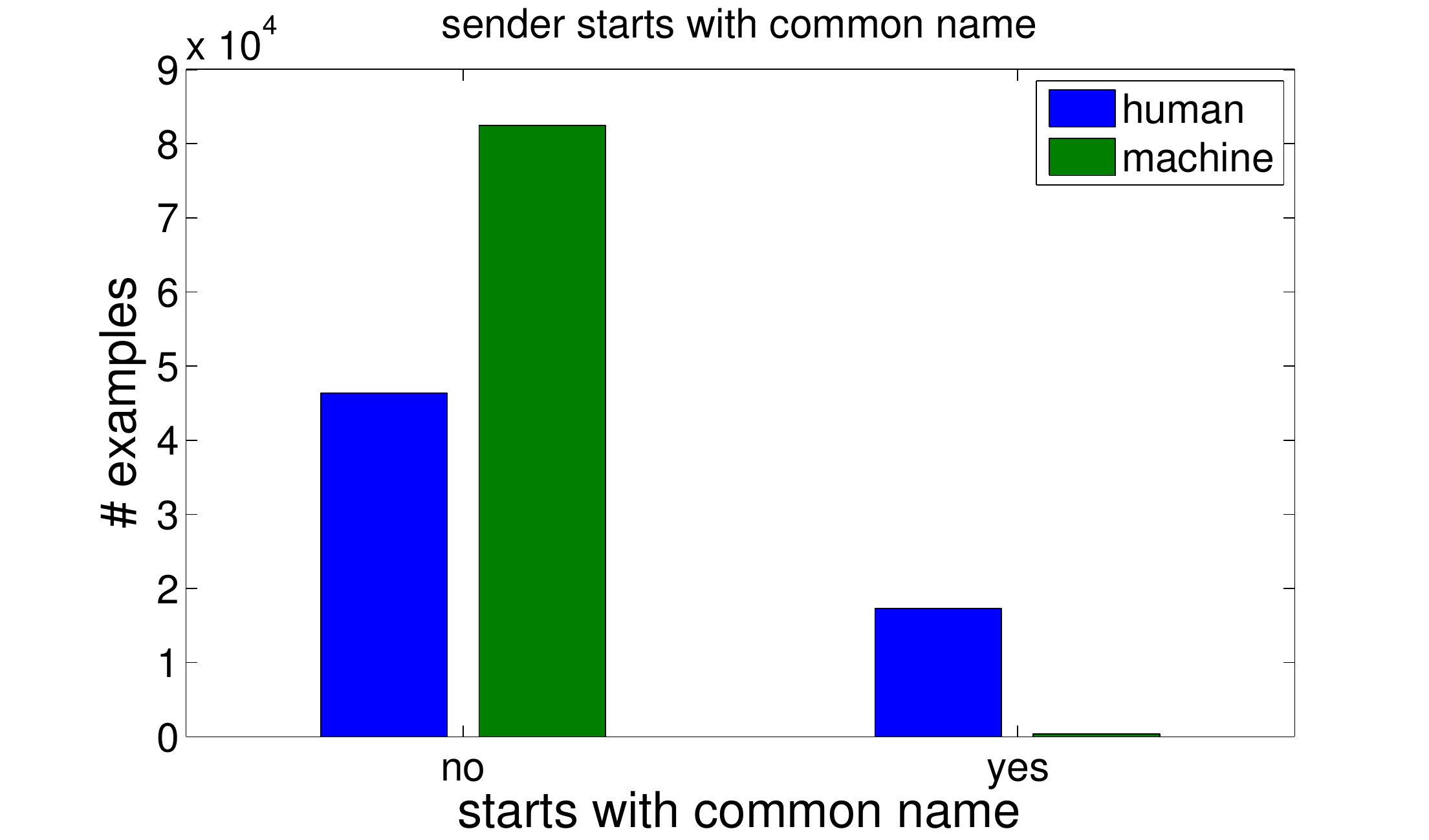}} 
\subfloat[{\small Burst-100 feature occurrence}]{\label{fig:humanBurst}\includegraphics[width=0.25\textwidth]{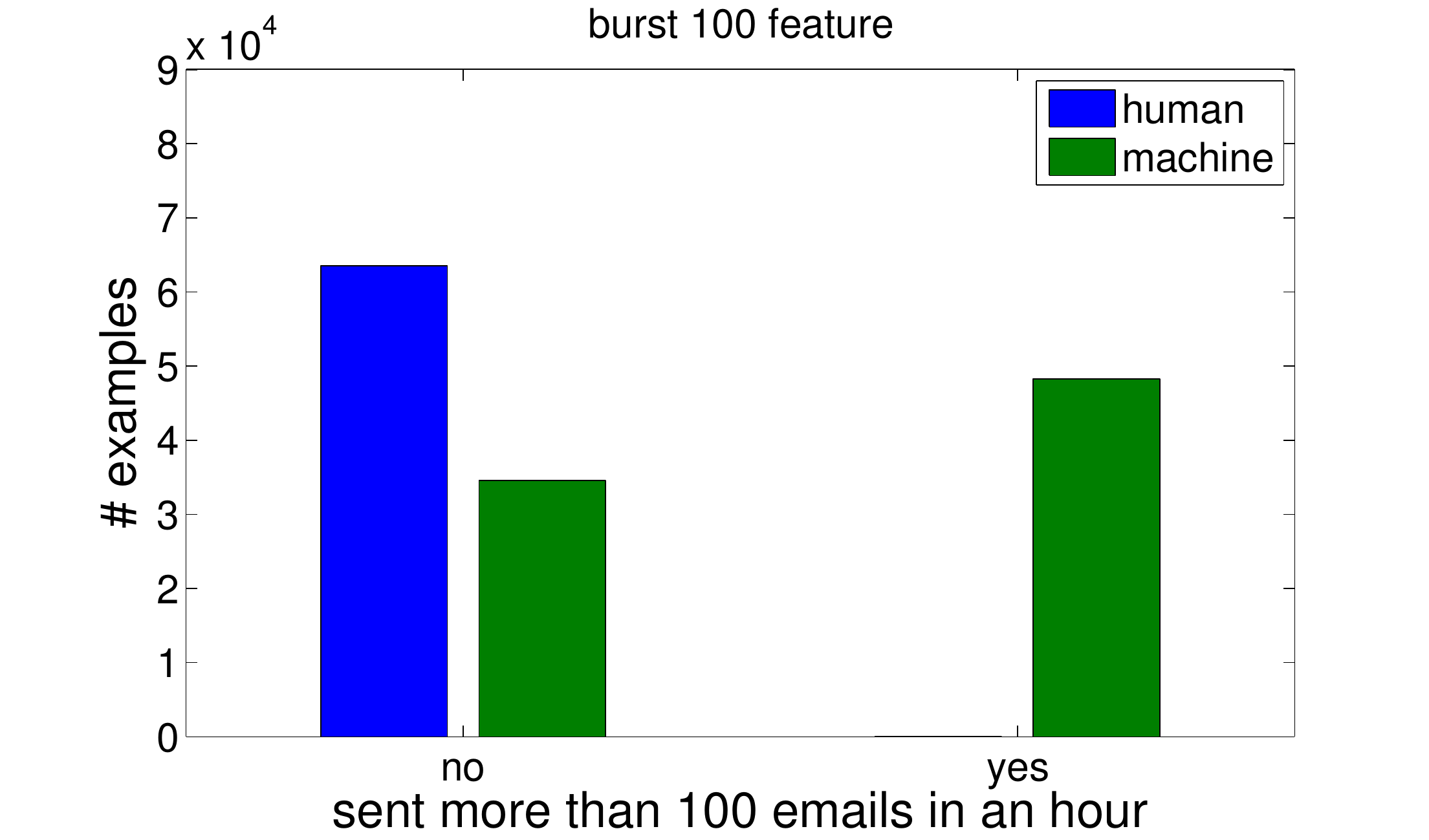}}
\caption{Statistics on human vs. machine data}
\vspace{-.2in}
\label{fig:resultsHuman}
\end{figure*}

\begin{figure*}
\centering
\subfloat[Shopping]{\label{fig:roc}\includegraphics[width=0.25\textwidth]{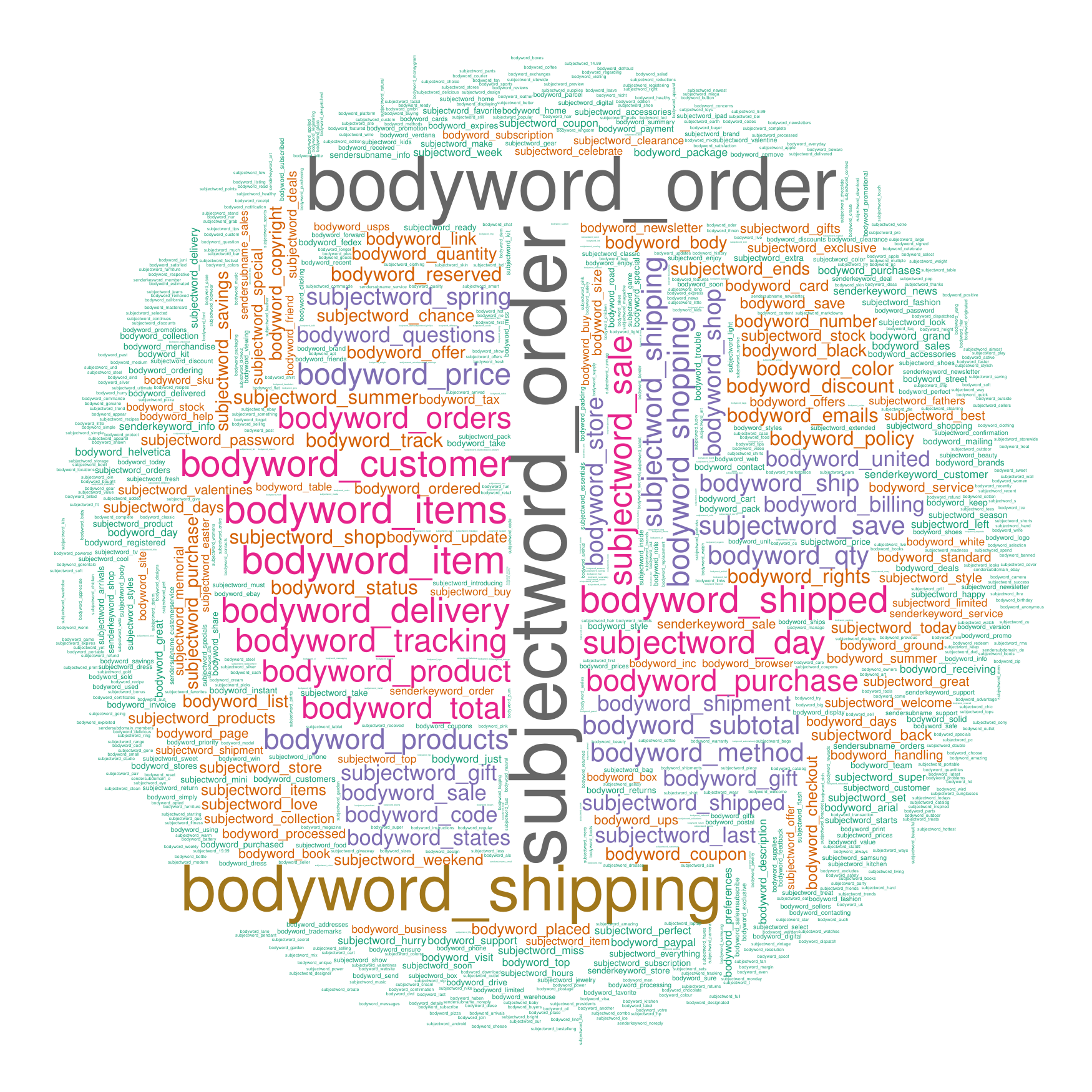}} 
\subfloat[Financial]{\label{fig:prcurve}\includegraphics[width=0.25\textwidth]{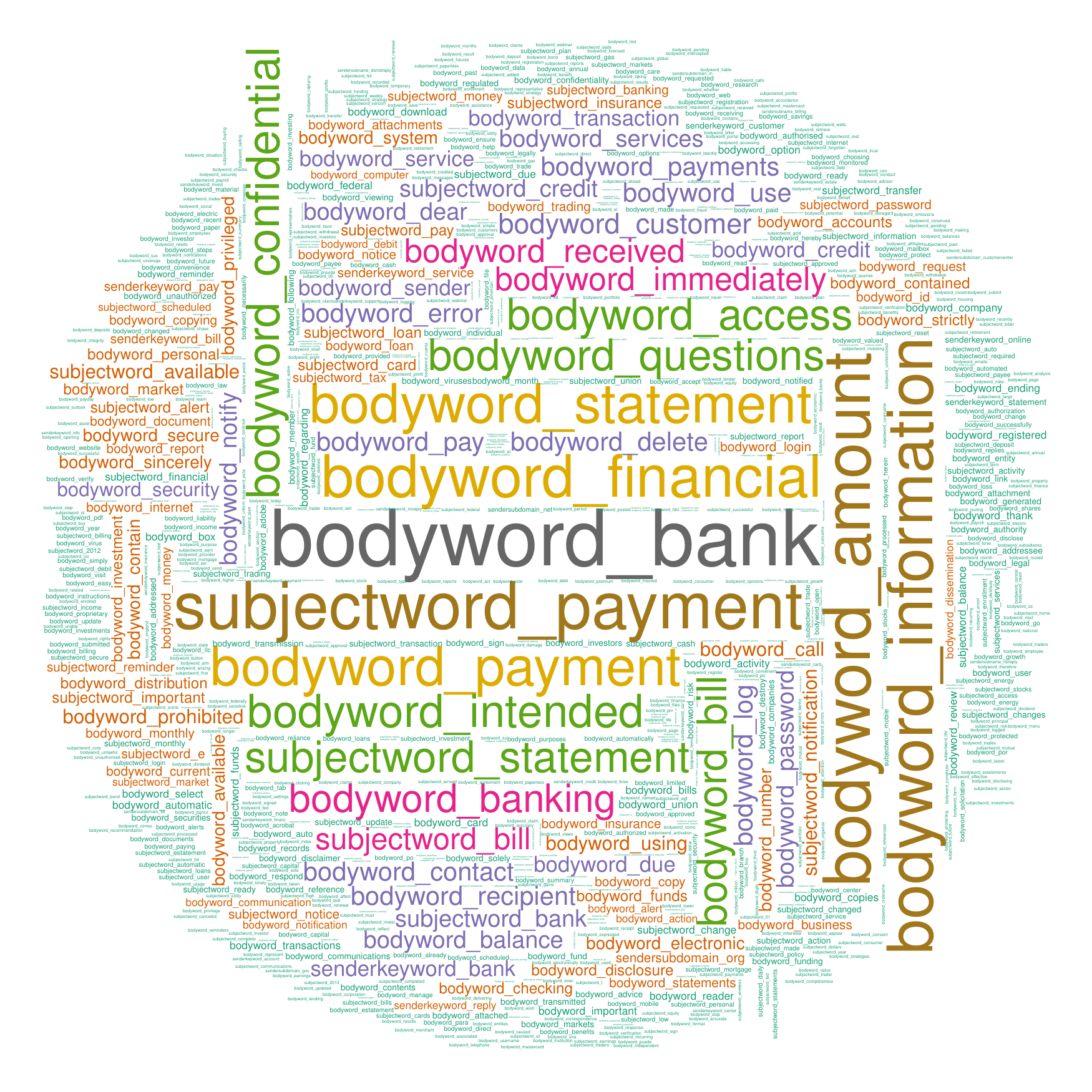}}
\subfloat[Travel]{\label{fig:roc}\includegraphics[width=0.25\textwidth]{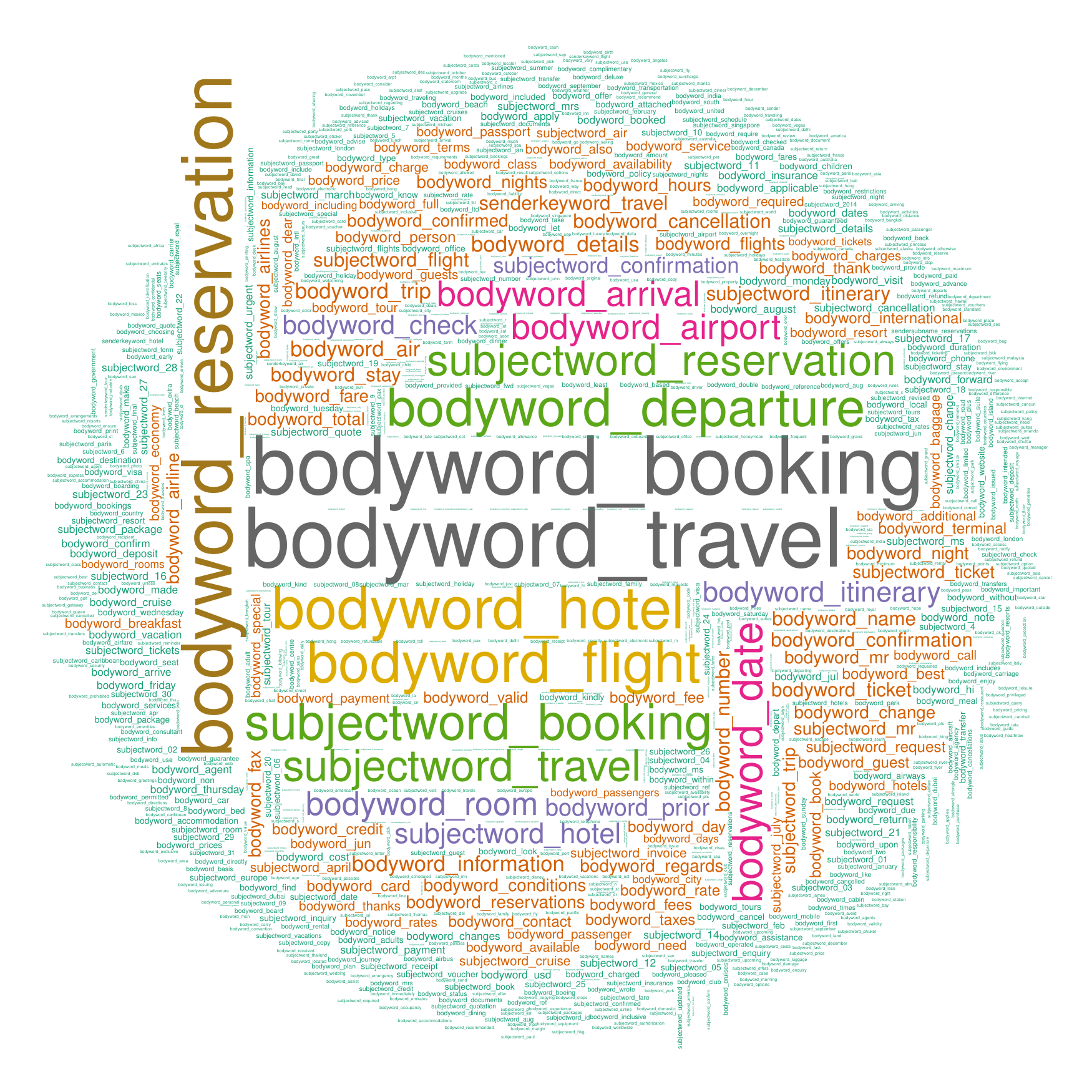}} 
\subfloat[Career]{\label{fig:prcurve}\includegraphics[width=0.25\textwidth]{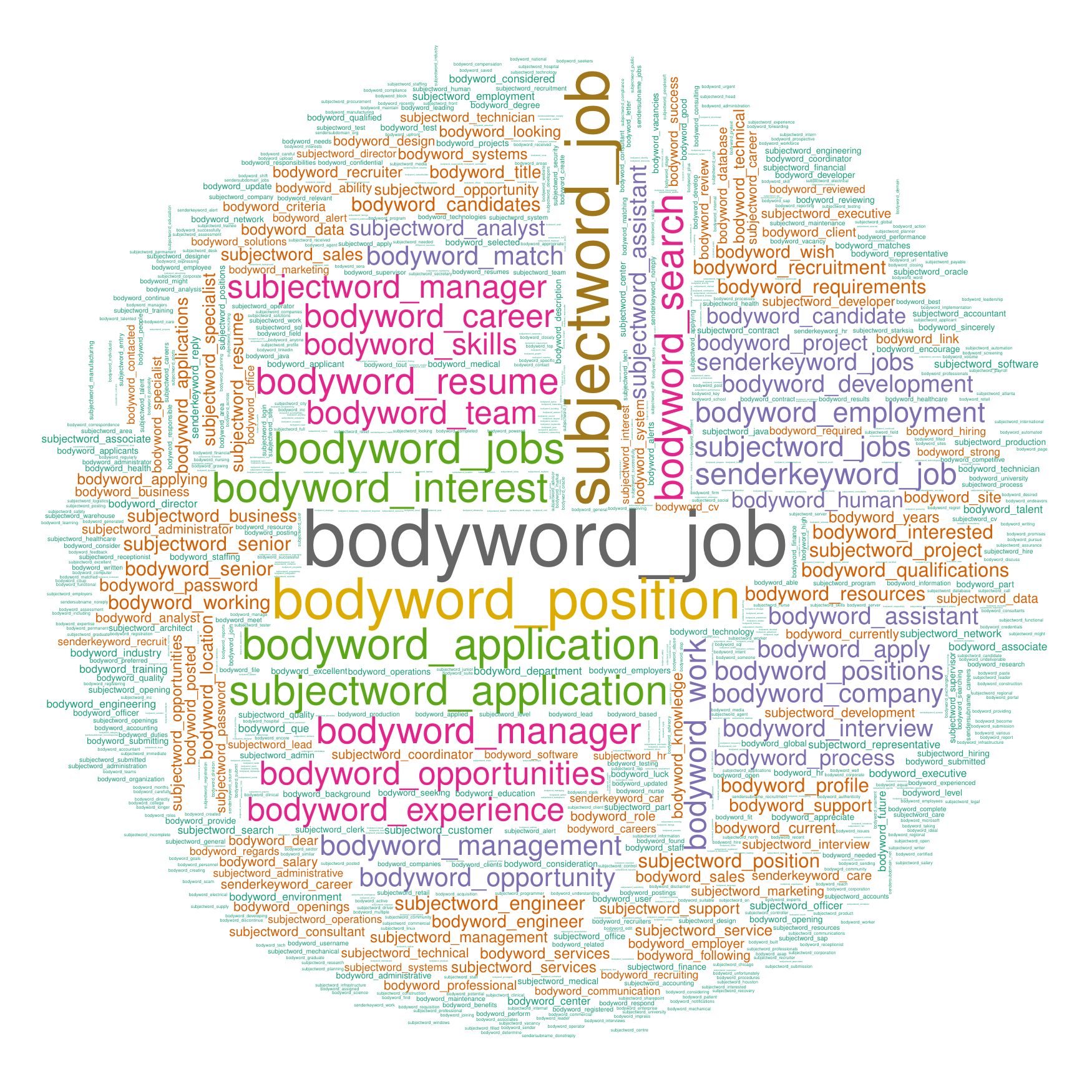}}
\caption{Feature clouds in each machine sub-category}
\label{fig:resultsMachine}
\end{figure*}

For the production system we trained $5$ sender-based classifiers for machine latent categories: {\em Shopping, Financial, Travel, Career} and {\em Social}, and $1$ sender-based machine vs. human classifier. This section covers the main experimental results of our email categorization study, demonstrating how well the category classifiers generalize on a holdout dataset.

\begin{figure}[t!]
\centering
  \includegraphics[width=0.33\textwidth]{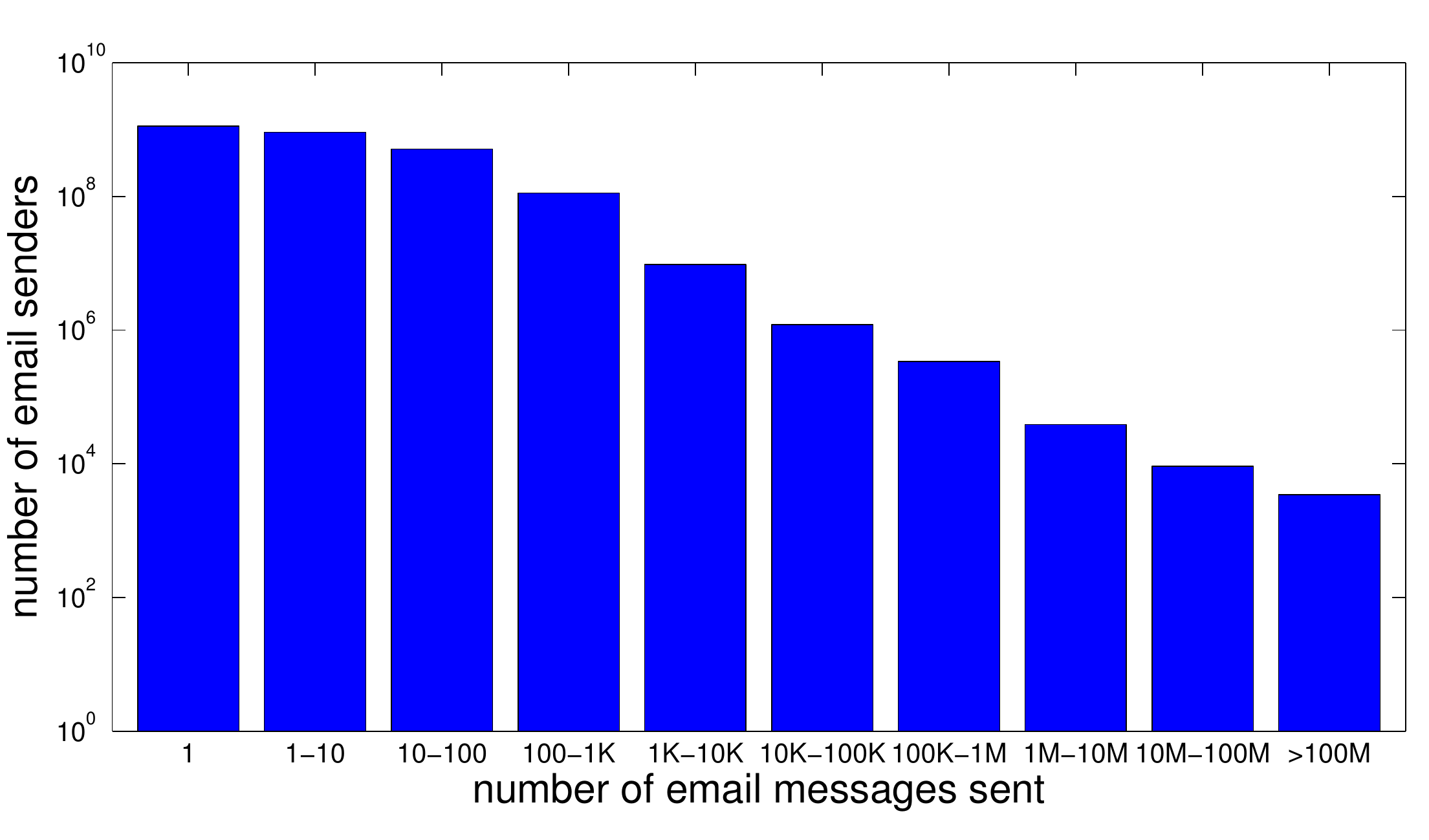}
  \captionof{figure}{Sender outbound traffic}
  \label{fig:Plowfig}
\end{figure}

We start by discussing the data distribution required for testing. The most intuitive sampling is uniform sampling, in which  each incoming message is assigned the same weight and the system is evaluated accordingly. In the machine-generated latent categories, our experiments did not consider this distribution as it would rate our system way too favorably: even an average system becomes indistinguishable from an excellent one due to the large volume of messages sent by top senders, as demonstrated in Figure~\ref{fig:Plowfig}.

Specifically,  a small number of senders are responsible for the majority of the email traffic, hence a system that avoids classifying the low-weight senders might achieve a good score, but still incur a bad user experience. One example for this phenomenon is in the social category. Here, Facebook contains only a handful of (canonized) senders\footnote{We define a canonized sender as a regular expression over senders, such that all senders matching it are essentially the same. In the context of Facebook, an example would be {\tt update+.*@facebookmail.com}} yet is in charge of a huge portion of the email traffic in the social category; as a matter of fact, setting a classifier for social that only outputs `True' for a message from Facebook would reach reasonable results when measured w.r.t the uniform distribution over email messages; clearly such a system would be horrible in terms of user's experience.
The distribution we use for testing is thus the uniform distribution over email senders, where canonized email senders e.g.\ \texttt{.*noreply@xyz.com} are viewed as a single sender. Due to skewed data and privacy issues, we did not rely on a manually labeled data set but rather on the data set $D_{v2}$ (ignoring the senders marked as human) described in Section~\ref{sec:training-set}. Although this set does not contain an actual uniform sample of senders, it covers senders whose categories are interesting to users, as their associated messages were manually assigned to folders. We mention that the machine sub-categories in the labeled set where chosen solely based on the folder data. For this reason, we excluded the folder features from the dataset. This was required to avoid over-fitting.

For the human-vs-machine classifier the training was performed on the set $D_{v2}$; however we used a different test set. The main reason is that unlike machine-labeled senders,  human-labeled senders are not based on folder data but rather on features that we do use in our classification process. For this reason, testing on $D_{v2}$ would provide unfair results. Instead, we sampled more than $600$ senders uniformly from the pool of senders, then from each sender we sampled $1$ email message thus creating a test set of more than $600$l messages. The samples were made from the email repository after  excluding easy-to-verify machine traffic such as senders whose outgoing traffic is larger than $100,000$  messages per month, or senders that were never replied to despite the fact that they sent over $1,000$ messages. This initial exclusion was necessary as human traffic is rather small compared to machine meaning that in order to get sufficiently many human examples without exclusion we would have required a large amount of manual labor.
We verified that a negligible amount of these excluded messages were written by a human by manually observing that out of $200$ uniformly drawn excluded examples none were human generated.

For the machine sub-category we randomly split $D_{v2}$ into a training set, used to train the models, and a test set, used to evaluate the models. Since a single domain may have multiple senders, they were all put in either the training or the test set. This way, we can test whether the model truly generalizes well outside of the domains it observed in training. The labeled dataset was split into $65\%$ for training and $35\%$ for testing purposes. The procedure was repeated $5$ times. The average numbers for the $5$ splits were reported.

In our experiments, as well as in the production system, we leveraged the MapReduce paradigm~\cite{dean2008mapreduce}, implemented in the Hadoop\footnote{\small \url{http://hadoop.apache.org/}} open source platform. All of our data processing, including aggregation of messages per sender, feature extraction, training and scoring, were done under Hadoop.  The models were trained in a single MapReduce job, where the mapper reads $D_{v2}$, and prints every data point $6$ times, once for each problem at hand, playing a role of either a positive or a negative example in the corresponding problems. The reducer trains a separate binary classification model:
\begin{displaymath}
f_j({\bf x}_i) \rightarrow y_i
\end{displaymath} for each problem
$j\in\{${\em Shopping, Financial, Travel, Career, Social, Human}$\}$.
We used a logistic regression model, where the feature vector ${\bf x}$ is parameterized using a weight vector ${\bf w}$, with one weight per modeled feature. Specifically, we utilized the highly scalable Vowpal Wabbit \footnote{\small \url{https://github.com/JohnLangford/vowpal_wabbit}} implementation of logistic regression~\cite{langford2009sparse}, which was found to work well in conjunction with MapReduce. In these experiments, the categorization was treated as a multi-class ``one-vs.-all'' problem. Therefore, an additional step of merging the predictions is required, where the final prediction is made based on the class with the highest probability. 

In Figure~\ref{fig:resultsROC}, we show the averaged results achieved on the test set when classifiers were trained using all feature types except folders. The figure shows the Receiver Operating Characteristic (ROC) curve that trades-off True Positive Rate (TPR) and False Positive Rate (FPR) as well as the Precision Recall (PR) curve that trades-off Precision and Recall. It can be observed that the models achieved desirable performance, with Precision and Recall numbers both in the $0.9$ range in most cases. 
Overall, the performance suggests that even with a very low FPR threshold, e.g. $0.5\%$, set for production purposes, large portions of true positives can be covered, i.e. , more than $85\%$ in most cases.

To evaluate the influence of the feature types, table~\ref{tab:resultsSelective} shows the Area Under the ROC curve (AUC) results for each class, when our model was trained on subsets of features: (1) content features (email body, subject, etc.), (2) address features (subname, subdomain, commercial keywords) and (3) behavioral features (number of messages, sent, received, etc.) without folders. It should be noted that temporal behavior features (burst features) were treated as behavioral features in this experiment. Even though there is an evident drop when compared to using all feature types, content and address feature subsets achieved competent AUC performance that around $0.9$. As expected, behavioral features showed good performance in distinguishing between human and machine senders, but could not on their own discriminate between different machine categories.

The second portion of Table~\ref{tab:resultsSelective} shows the performance when some features were removed. Specifically, we were interested in the performance drop when burst features, and body words features were removed. 

Figure~\ref{fig:resultsHuman} gives some insights on the human vs. machine classification. Figure~\ref{fig:humanRead} shows the histogram of the read ratio $R_{rr}$ for human and machine senders, where $R_{rr}$ is defined as $R_{rr}=\frac{\#\text{read messages}}{\#\text{sent messages}}$.

A clear difference in the distribution can be observed. Furthermore, Figures ~\ref{fig:humanUn} and ~\ref{fig:humanCommon} show to what extent certain binary features are observed in the human examples versus the machine examples. It can be observed that when the word {\tt unsubscribe} is present in the message body, it is much more likely to be associated with a machine sender. Also, when common first names, such as ``michael'', ``susan'', etc., are present in the email address, they are much more likely to be associated with a human sender. Finally, Figure~\ref{fig:humanBurst} shows the presence of ``burst 100'' feature, i.e. indicator of whether or not the sender sent more than $100$ messages per hour on some occasion. It can be observed that the feature is present in more than half machine senders and in very few human senders. Other interesting facts that were not depicted in the figures are that: 1) $34.56\%$ of machine senders send messages with average subject character count higher than $30$, while only $5.88\%$ of human senders do the same; 2) $77.55\%$ of machine senders send messages with average body character count higher than $300$, while only $1\%$ of human senders do the same; and 3) $41.09\%$ of machine senders send messages with more than $3$ urls in the body on average, while only $2.24\%$ of human senders do the same.

To get more insight into the latent category models, we generated feature clouds for $j\in\{$ {\em Shopping, Financial, Travel, Career}$\}$ as illustrated in Figure~\ref{fig:resultsMachine}. Features of higher ``importance'' are printed using larger fonts. In Logistic Regression, large positive feature weights ${\bf w}$ are associated with a higher likelihood of user belonging to class $j$. Therefore, to calculate the feature importance scores with respect to class $j$ we use the following procedure. First, we isolate only the features that have a positive weight in $j$-th class model ${\bf w}_j$. Next, we calculate the score for each of those features as $score(k)=N_k (\frac{N_k^+/N_k}{N^+/N} - 1)$, where $N$ is the total number of examples, $N^+$ is the total number of class $j$ examples, $N_k$ is the number of examples that have feature $k$ and $N_k^+$ is the number of class $j$ examples that have feature $k$. Finally, we generate a word cloud using the $1,000$ features with the highest score. By examining the figures we can conclude that the main concepts of each class are very well captured using indicative body and subject words, as well as the address substrings.
 
\section{Significance of Results}
\label{sec:results}
In this section, we discuss the significance and potential impact of
our mail categorization system. We estimated the
coverage of latent categories in two different manners. First, we
measured overall email traffic coverage by estimating the percentage
of email messages that are mapped into one of our categories. Then, we
measured how these categories cover mail search by manually
categorizing a sample of mail search queries.\comment{ , i.e.\ search
  expressions used in the search textbox within the email client, and
  estimated the percentage of search queries related to each latent
  topic. } The results are given in Table~\ref{tab:traffic_results}.
\begin{table}[h]
\centering
{\scriptsize
  \caption{Coverage per latent category}
  \vspace{-.1in}
    \begin{tabular}{l | l | l }
      {\bf category} & {\bf email traffic (\%)} & {\bf search query (\%)}  \\
      \hline
human & $4.71$ & $25.12$ \\
social & $33.49$  & $2.7$ \\
shopping &  $22.28$  & $22.73$ \\
travel & $3.00$  & $6.48$\\
financial & $5.55$ & $11.64$ \\
career & $4.11$& $3.62$ \\
\hline 
Total & $73.14$& $72.29$ \\
    \end{tabular}
  \label{tab:traffic_results}
}
\end{table}

For email traffic coverage, we ran our classification system over real
email traffic and counted the number of messages classified in each
category. The percentages listed in the email traffic column in
Table~\ref{tab:traffic_results} represent the number of messages
classified as topic X divided by the total number of messages.  For
the search query coverage,  a team of professional editors manually labeled
2,500 common queries. To ensure privacy, we chose only queries that
were used by several users, and in particular avoided queries that are
unique to a single user. The percentages listed in the search query
column in Table~\ref{tab:traffic_results} represent the number of
queries classified as topic X divided by the total number of
queries. Table~\ref{tab:query_mini_analysis} gives a description of the typical queries for each category.

\begin{table}[h]
\centering
{\scriptsize
  \caption{Typical queries related to latent categories}
  \vspace{-.1in}
    \begin{tabular}{l | l  }
      {\bf category} & {\bf typical queries}  \\
      \hline
human & person's name, terms from a conversation \\
& with another person \\

social & social network name such as twitter, pinterest, etc. \\
shopping &  retailer name, product type or product name \\
travel & travel airline, hotel name, destination, travel \\
&  keywords (such as travel)\\

financial & company name (e.g., bank or insurance company), \\
& keywords such as tax, loan or insurance \\

career & company name (e.g., careerbuilder, ziprecruiter, etc), \\
& keywords such as job, contract, etc. \\

    \end{tabular}
  \label{tab:query_mini_analysis}
}
\end{table}

Table~\ref{tab:traffic_results} clearly illustrates the fact that
coverage by email traffic and coverage by search query are very
different in nature. One notable difference is  the human
category that attracts (somewhat unsurprisingly, given our previous
comments on the dominance of machine-generated traffic) a disproportionate
amount of searches as compared to its relatively small email volume. The finance
category is similar although the ratio between the coverages is not as
large as in the human category. An opposite example is that of social
networks. Despite their huge coverage of the email traffic, the
coverage in terms of search queries is much lower. Indeed when
considering the type of messages sent by social sites, these are
mostly updates and summaries of the recent events, hence are typically
not messages that will be read more than once, thus will not be
searched for. Overall, the coverage of the latent categories
both in terms of email traffic and search queries is larger than
70\%. These figures confirm the fact that latent
categories could be used not only for browsing but probably even more
for searching, and thus answer the needs of the two traditional
discovery paradigms~\cite{Bowman.94} since the early days of the Web.

\section{Conclusions and future work}
\label{sec:conclusion}
We presented here a Web-scale categorization approach that combines offline learning and online classification components.  One of our key contributions is the identification of categories common to all users. We discovered latent categories by conducting a large-scale analysis of user folder data, which highligted the distinction between human and machine-generated email. Our categories cover more than 70\% of both email traffic and email search queries. Our classification mechanism achieved precision and recall rates close to 90\%. These results are achieved via an extremely scalable online system that assigns a category to incoming messages delivered to any user, including those who never defined a folder. We believe that this study shows a great deal of promises for this domain, as it demonstrates that email classification can be applied in production systems of the scale of Web mail. 

Discussing how categories should be exposed to users, if at all, to users is out of the scope of this work but is clearly a critical challenge. Following the traditional Web discovery paradigms, they could be surfaced by explicit categories and search facets (for browsing) or behind the scenes, as an additional search signal. We hope this work will encourage other researchers to build upon these categories in order to explore new email discovery paradigms. On the backend side, which remains our focus, we plan to investigate whether sub-types within latent categories could be discovered, e.g., {\em Travel promotions} under {\em Travel}.  We also intend to explore the ``sender cold-start'' issue, which occurs when a new sender appears, and its associated messages are too rare for us to conduct appropriate learning. As new senders keep appearing, we will need to devise appropriate methods to handle new senders in order to maintain quality.

\section*{Acknowledgements}
We are grateful to Andrei Broder for inspiring this work. Yehuda Koren and Roman Sandler spent a huge number of hours manipulating thousands of folders. We owe a great deal to Edo Liberty, who discovered the power of machine-generated email, and Nemanja Djuric who conducted LDA experiments during his internship at Yahoo. Finally, this work would not be possible without the constant support of the Mail engineering and product teams at Yahoo.

\bibliographystyle{plain}
\bibliography{cikm14_refs}

\begin{thebibliography}{10}

\bibitem{ailon.2013}
Nir Ailon, Zohar~S. Karnin, Edo Liberty, and Yoelle Maarek.
\newblock Threading machine generated email.
\newblock In {\em Proceedings of WSDM'2013}, pages 405--414, New York, NY, USA,
  2013. ACM.

\bibitem{alberts2012email}
Inge Alberts and Dominic Forest.
\newblock Email pragmatics and automatic classification: A study in the
  organizational context.
\newblock {\em JASIST}, 63(5):904--922, 2012.

\bibitem{balter2000keystroke}
Olle B{\"a}lter.
\newblock Keystroke level analysis of email message organization.
\newblock In {\em Proceedings of CHI'2000}, pages 105--112. ACM, 2000.

\bibitem{blei2003latent}
David~M Blei, Andrew~Y Ng, and Michael~I Jordan.
\newblock Latent dirichlet allocation.
\newblock {\em JMLR}, 3:993--1022, 2003.

\bibitem{blum1998combining}
Avrim Blum and Tom Mitchell.
\newblock Combining labeled and unlabeled data with co-training.
\newblock In {\em Proceedings of COLT'1998}, pages 92--100. ACM, 1998.

\bibitem{Bowman.94}
C.~M. Bowman, P.~B. Danzig, U.~Manber, and M.~F. Schwartz.
\newblock Scalable internet: Resource discovery.
\newblock {\em Communications of the ACM}, 37(8), August 1994.

\bibitem{brutlag2000challenges}
Jake~D Brutlag and Christopher Meek.
\newblock Challenges of the email domain for text classification.
\newblock In {\em Proceedings of ICML'2000}, pages 103--110, 2000.

\bibitem{dean2008mapreduce}
Jeffrey Dean and Sanjay Ghemawat.
\newblock Mapreduce: simplified data processing on large clusters.
\newblock {\em Communications of the ACM}, 51(1):107--113, 2008.

\bibitem{hoffman2010online}
Matthew Hoffman, Francis~R Bach, and David~M Blei.
\newblock {Online learning for Latent Dirichlet Allocation}.
\newblock In {\em {Advances in Neural Information Processing Systems}}, pages
  856--864, 2010.

\bibitem{kiritchenko2011email}
Svetlana Kiritchenko and Stan Matwin.
\newblock Email classification with co-training.
\newblock In {\em Proceedings of the 2011 Conference of the Center for Advanced
  Studies on Collaborative Research}, pages 301--312, 2011.

\bibitem{SMTP}
J.~Klensin.
\newblock Simple mail transfer protocol, rfc 2821, April 2001.

\bibitem{klimt2004enron}
Bryan Klimt and Yiming Yang.
\newblock The enron corpus: A new dataset for email classification research.
\newblock In {\em Proceedings of ECML'2004}, pages 217--226. 2004.

\bibitem{koren2011automatically}
Yehuda Koren, Edo Liberty, Yoelle Maarek, and Roman Sandler.
\newblock Automatically tagging email by leveraging other users' folders.
\newblock In {\em Proceedings of KDD'2011}, pages 913--921. ACM, 2011.

\bibitem{langford2009sparse}
John Langford, Lihong Li, and Tong Zhang.
\newblock Sparse online learning via truncated gradient.
\newblock {\em JMLR}, 10:777--801, 2009.

\bibitem{provost1999naive}
Jefferson Provost.
\newblock Na{\i}ve-bayes vs. rule-learning in classification of email.
\newblock {\em University of Texas at Austin}, 1999.

\bibitem{sebastiani2002}
Fabrizio Sebastiani.
\newblock Machine learning in automated text categorization.
\newblock {\em ACM Computing Surveys}, 34(1), March 2002.

\end{thebibliography}

\end{document}